\documentclass[conference]{IEEEtran}
\usepackage{amsmath}
\usepackage{color}
\usepackage{amssymb}
\usepackage{enumerate}
\usepackage{helvet}
\usepackage{courier}
\usepackage{diagbox}
\usepackage{amsmath}
\usepackage{balance}
\usepackage{amssymb}
\usepackage{makecell}

\usepackage{multirow}
\usepackage{graphicx}
\usepackage{subfig}
\usepackage{algorithm}
\usepackage{algorithmic}
\usepackage{amsmath}
\usepackage{graphics}
\usepackage{epstopdf}
\usepackage{enumerate}\usepackage[colorinlistoftodos]{todonotes}
\usepackage{mathrsfs,amsmath}
\def\ie{{\em i.e.}}
\def\eg{{\em e.g.}}

\begin{document}
\title{Camera Fingerprint: A New Perspective for\\
Identifying User's Identity}

\author{\IEEEauthorblockN{Xiang Jiang\IEEEauthorrefmark{1}, Shikui Wei\IEEEauthorrefmark{1}, Ruizhen Zhao\IEEEauthorrefmark{1}, Yao Zhao\IEEEauthorrefmark{1}, Xindong Wu\IEEEauthorrefmark{2}}
\\
\IEEEauthorblockA{\IEEEauthorrefmark{1}Institute of Information Science, Beijing Jiaotong University, China \\ 
\IEEEauthorrefmark{1}Beijing Key Laboratory of Advanced Information Science and Network Technology, Beijing 100044, China.\\
\IEEEauthorrefmark{2}Department of Computer Science, University of Vermont.\\
Email: \{xiangj, shkwei, rzhzhao, yzhao\}@bjtu.edu.cn, xwu@hfut.edu.cn}
}
\maketitle

\begin{abstract}

Identifying user's identity is a key problem in many data mining applications, such as product recommendation, customized content delivery and criminal identification. Given a set of accounts from the same or different social network platforms, user identification attempts to identify all accounts belonging to the same person. A commonly used solution is to build the relationship among different accounts by exploring their collective patterns, e.g., user profile, writing style, similar comments. However, this kind of method doesn't work well in many practical scenarios, since the information posted explicitly by users may be false due to various reasons. In this paper, we re-inspect the user identification problem from a novel perspective, \ie, identifying user's identity by matching his/her cameras. The underlying assumption is that multiple accounts belonging to the same person contain the same or similar camera fingerprint information. The proposed framework, called \textit{User Camera Identification} (\textbf{UCI}), is based on camera fingerprints, which takes fully into account the problems of multiple cameras and reposting behaviors.
To facilitate the assessment of the proposed framework, we generate a new benchmark called  \textbf{UID-BJTU}. Extensive experiments on this benchmark show that the proposed framework is quite effective for identifying the user's identity, especially on the case with multiple cameras and reposted images.

\end{abstract}

\IEEEpeerreviewmaketitle

\section{Introduction}
Online social networks are becoming more and more popular. According the related report \cite{Kemp2015},  the number of registered accounts on online social networks has exceeded 3 billion, and it continues to grow routinely. However, each account doesn't exclusively correspond to a person. Generally, the average number of social media accounts is about 6 for an individual. The main reason lies in that people usually register multiple accounts on several platforms for different social needs. For example, people always communicate with workmates on LinkIn, maintain the friend contact on Facebook, and follow the celebrities on Twitter, etc. Therefore, analyzing social behaviors from a single account is partial for characterizing a user. Towards
constructing a relatively complete user profile, we need to identify all the accounts belonging to the same individual.

In recent years, the studies of user identification have drawn more attention, since it is beneficial to many practical applications \cite{Tang, Wei2016, Li2016, Li2016a, Fang2015}. First, we can build a more complete user profile by fully exploiting behavior patterns converged from multiple accounts, which can be used to provide personalized network services, such as customized content delivery or friend recommendation. In addition, having comprehensive view of users would make it easier to determine user's identity in real world, which is very helpful to fight against the Internet criminal, terrorism, Internet porn and other social problems \cite{Satta2014,Interpreting2014,Cross-Platform2016,Personalized2014}.

\begin{figure}[t]
	\centering
	\includegraphics[height=5cm]{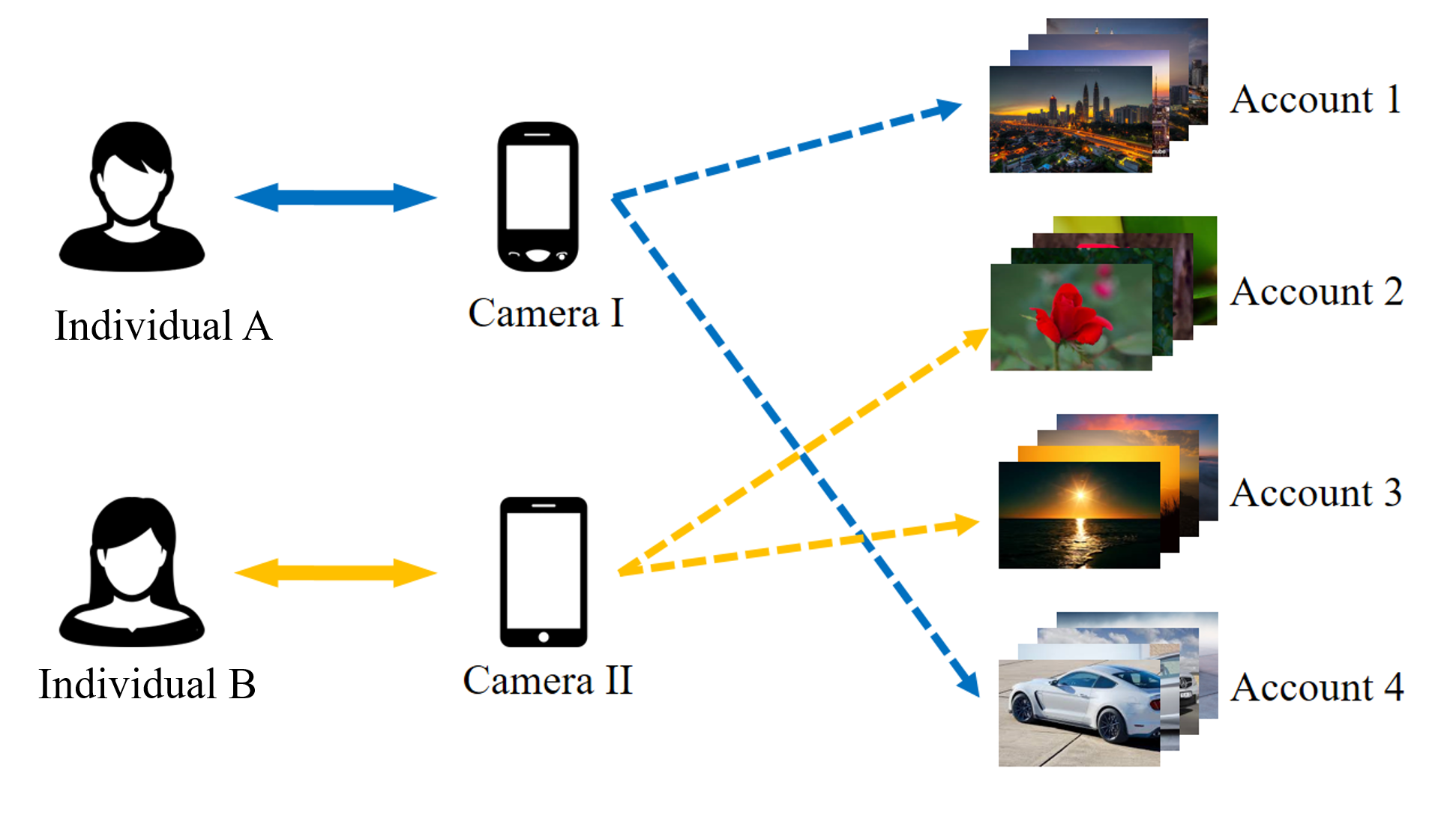}
	\caption{\small
		The basic idea of our approach. Given four accounts, accounts 1 and 4 are asserted to belong to the same individual A, since all images in accounts 1 and 4 are captured by the same camera I. 
	} 
	\label{fig:fig1}
\end{figure}
%

The core problem of user identification is how to correctly estimate similarity between two different accounts by employing some discriminative user patterns. However, the problem is very challenging since it is difficult to discover a kind of universal pattern for an individual with diverse accounts. Traditionally, user patterns are extracted from either the user's public profiles ~(\eg, username, geography location and E-mail address) \cite{Iofciu2011} or activity characters~(\eg, linguistic stylistics, writing styles)\cite{Narayanan2012, Zafarani2013}. In fact, these patterns are not always reliable enough to identify user's real identity, since all of them can be easily faked. Therefore, discovering a universal and reliable user pattern is the fundamental step of user identification.

%
%

\begin{figure*}[t]
	\centering
	\subfloat[Multiple camera problem]
	{\label{fig:subfig:a}
		\includegraphics[width=6cm]{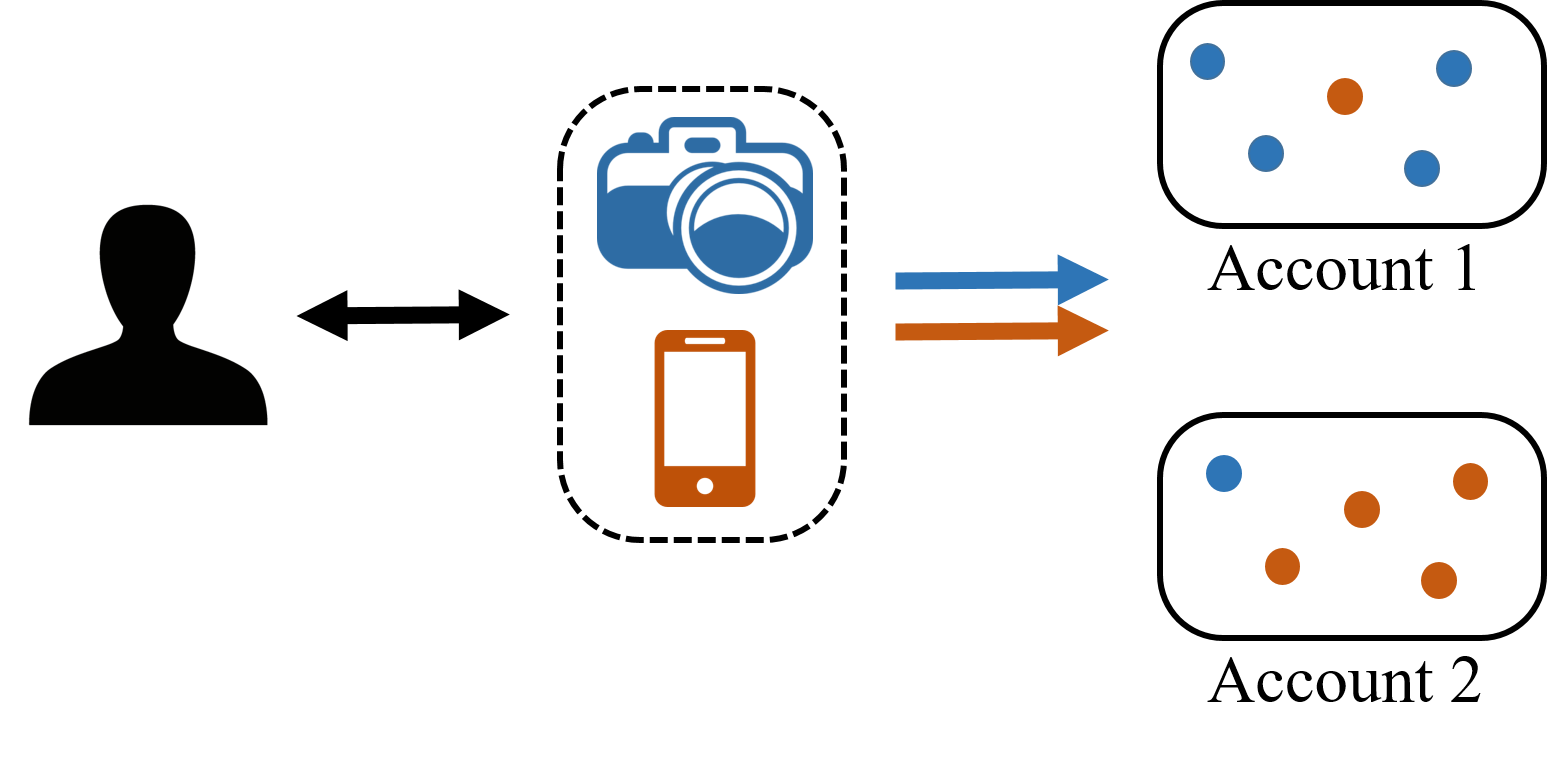}}\qquad
	\subfloat[Reposted image problem]
	{\label{fig:subfig:b}
		\includegraphics[width=6cm]{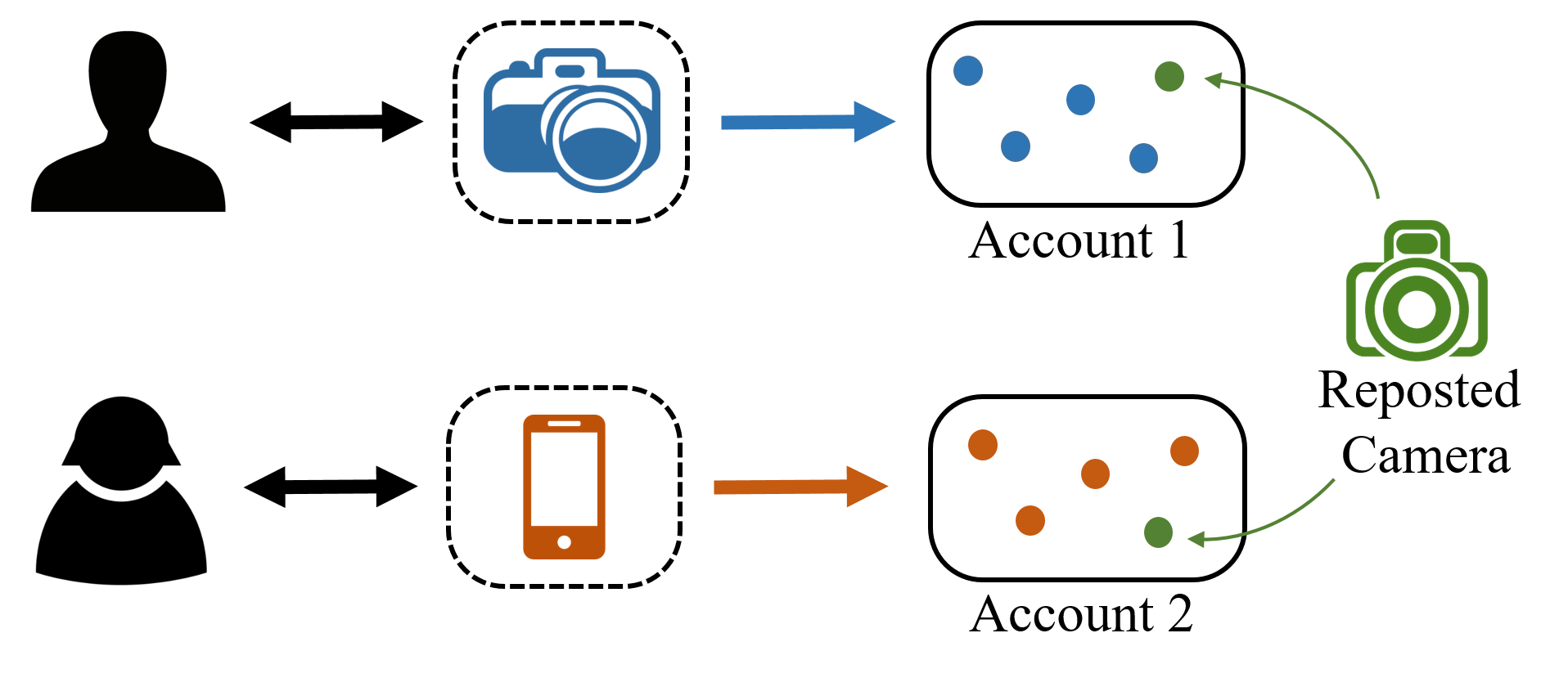}}\\
	\subfloat[Single camera sharing problem]
	{\label{fig:subfig:c}
		\includegraphics[width=6cm]{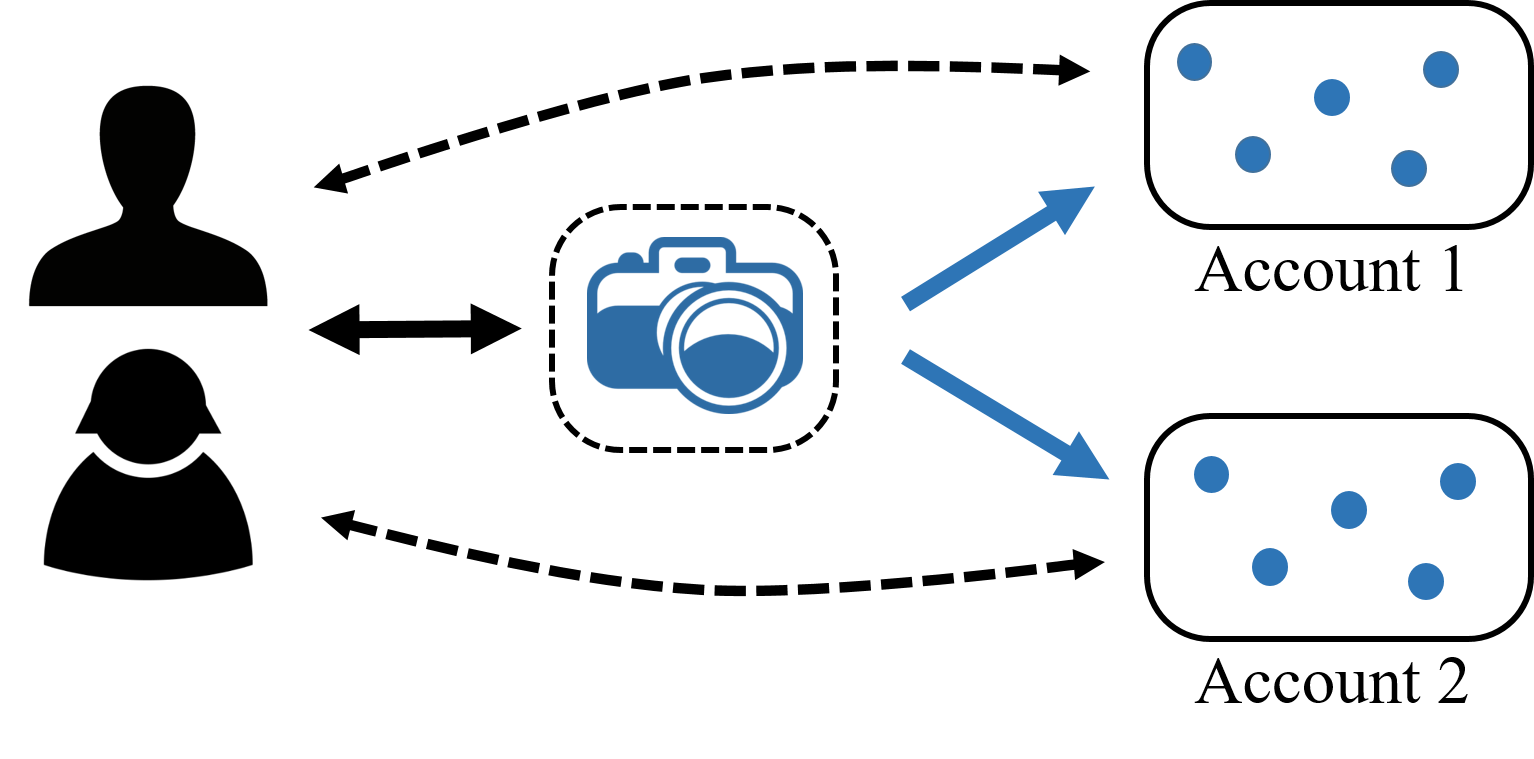}}\qquad
	\subfloat[Multiple camera sharing problem]
	{\label{fig:subfig:d}
		\includegraphics[width=6cm]{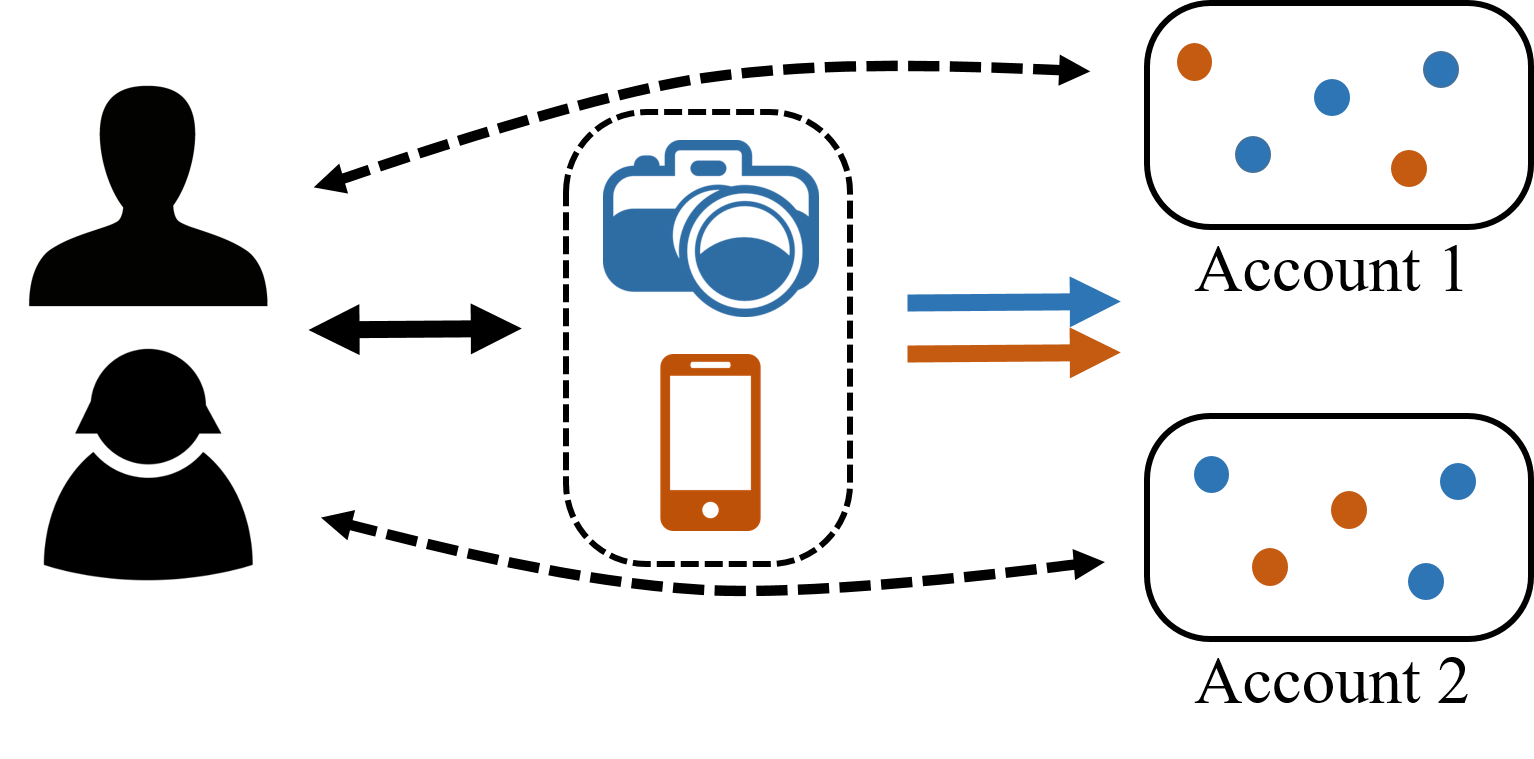}}
	\caption{\small
		Four typical challenges in camera fingerprint based user identification framework.
	} 
	\label{fig1_2}
\end{figure*}

In this paper, we aim to re-inspect the user identification problem from a totally different perspective, \ie, identifying users by matching their photography devices. The basis idea is based on a generally existing finding that most of photos published in different accounts by the same person are captured by several frequently used photography devices. That is, different accounts belonging to the same person include the same implicit fingerprint information of these cameras. Camera fingerprint is a kind of discriminative and reliable feature extracted from images, which is traditionally used for digital camera identification. Totally different from the explicitly published information, camera fingerprint is mainly dependent on the light sensitivity of different camera sensors, which is unique for each camera and difficult to forge \cite{Lukas2006}. The basic idea is illustrated in Fig.\ref{fig:fig1}, where accounts 1 and 4 are asserted to belong to the individual A since all images in these accounts are from the same camera source.


Intuitively, we can individually extract camera fingerprints for all accounts and identify users by directly estimating the similarity of camera fingerprints. However, the online social networks are far more complex than what we think, and the relationship among individuals, accounts and cameras are also complex and uncertain. To make our exposition more clear, we first illustrate some difficulties when identifying users from the new perspective.


\begin{itemize}
\item
Multiple camera problem. As shown in Fig.\ref{fig:subfig:a}, a person commonly owns more than one camera. That is, it is possible that the images in an account are from more than one camera source. If we estimate only a single camera fingerprint for the account, both its discriminative capability and reliability will inevitably decreased. Therefore, it is vital to distinguish the camera sources of images in the same account in an explicit or implicit manner.

\item 
Reposted image problem. As shown in Fig.\ref{fig:subfig:b}, two different users may repost some popular images from the same camera source (called Reposted camera).
If the reposted images are not removed before estimating camera fingerprints, two different users may not be clearly distinguished by the confused user patterns. Therefore, it is necessary to eliminate the effect of reposted images.

\item 
Single camera sharing problem. As shown in Fig.\ref{fig:subfig:c}, two different individuals share the same camera source. In this case, it is extremely difficult to distinguish two users by using only camera fingerprints. 

\item 
Multiple camera sharing problem. As shown in Fig.\ref{fig:subfig:d}, multiple individuals share multiple camera sources. In this case, the relationship between users and cameras is more complex, and some new strategies should be taken into account.

\end{itemize}

However, it is very challenging to address all these problems in a single scheme. Therefore, we focus mainly on first two problems in this paper and propose a camera fingerprint based user identification framework, called \textit{User Camera Identification} (UCI) to address them. To the best of our knowledge, it is the first time to re-inspect the social network reconciliation problem from the perspective of camera fingerprint. The main contributions can be summarized as follows:

\begin{itemize}
\item  A new perspective is proposed to tackle the user identification problem. Totally different from previous methods, the proposed approach explores camera fingerprints to identify user's identity, which is more reliable and difficult to forge. In addition, the new perspective makes it possible to meet the requirements of many practical applications, such as criminal detection.

\item 
A novel estimation approach of camera fingerprints is proposed to incrementally extract multiple camera fingerprints in an account. Different from existing methods, the proposed approach can significantly deal with the confused problem caused by multiple cameras and reposed images, which  is beneficial to both user identification and camera identification areas.

\item 
A new dataset is constructed to benchmark camera fingerprint based user identification framework. We will release this dataset so as to provide a benchmark for evaluating new approaches.
\end{itemize} 


\section{Problem Statement}
Given a user $U_i$, its image set is denoted to  $\mathcal{I}_{i}$ $=\{I_{i;1},...,I_{i;j},...,I_{i;m}\}$, where $I_{i;j}$ represents the $j^{th}$ image of user $U_i$. The core problem is how to determine whether users $U_x$ and $U_y$ belong to a single individual or not. In our work, we attempt to employ camera fingerprint to address this problem. Therefore, the problem is converted to extract camera fingerprints of users and measuring the similarity of any two users' camera fingerprints.

Before defining our problem, we first discuss the process of camera fingerprint extraction. Based on \cite{Lukas2006, Chen2008}, given a set of images $\mathcal{I}_i$ (captured by the same photography device), we first extract a residual noise $R_{i;j}$ for any image $I_{i;j}$ as follows 

\begin{equation}
\label{eq.2}
R_{i;j}=I_{i;j}-F(I_{i;j}),
\end{equation}
where $F(\cdot)$ is a wavelet denoising filter. Then, a camera fingerprint $S_{i}$ is achieved by averaging all images' residual noises as follows

\begin{equation}
\label{equ:3}
S_{i}=\sum_{j=1}^{m}\frac{I_{i;j}\cdot R_{i;j}}{I_{i;j}^2}. 
\end{equation}

For two camera fingerprints $S_{x}$ and $S_{y}$ of users $U_x$ and $U_y$, their similarity can be measured by

\begin{equation}
\label{equ:4}
corr(S_x,S_y)=\frac{(S_x-\bar{S_x})\cdot(S_y-\bar{S_y})}{\|S_x-\bar{S_x}\|_2\cdot\|S_y-\bar{S_y}\|_2},
\end{equation}
where $\bar{S_x}$ and $\bar{S_y}$ are the means of $S_x$ and $S_y$, respectively. The correlation value $corr(\cdot)$ can be taken as the similarity score of these two cameras.

\begin{figure}[t]
\centering
\includegraphics[height=5cm]{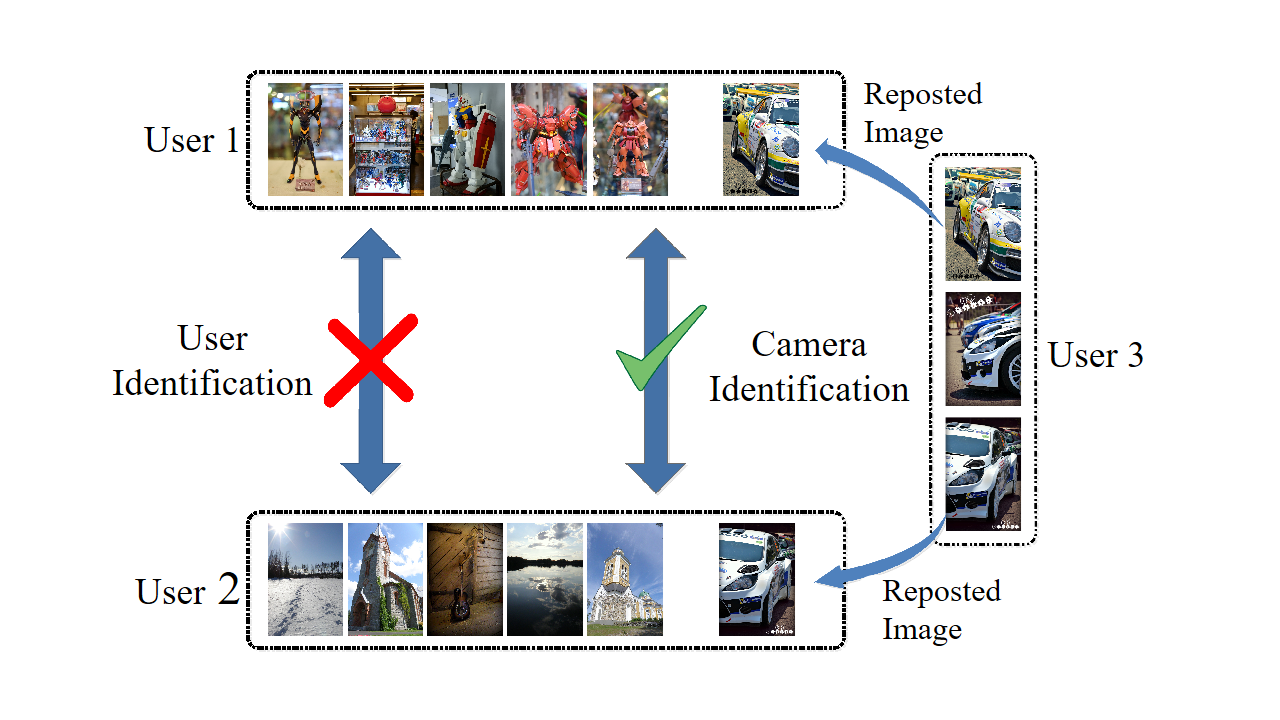}
\caption{\small
Confusion problem of reposted images. Users 1 and 2 don't belong to the same individual, but both of them repost the images from user 3. Without tackling this issue, users 1 and 2 will be incorrectly asserted to belong to the same individual, since reposted images are captured by the same camera.
} 
\label{fig:reposting}
\end{figure}

\begin{figure*}[t]
\centering
\subfloat[]
{\label{fig:subfig:i}
\includegraphics[width=7cm]{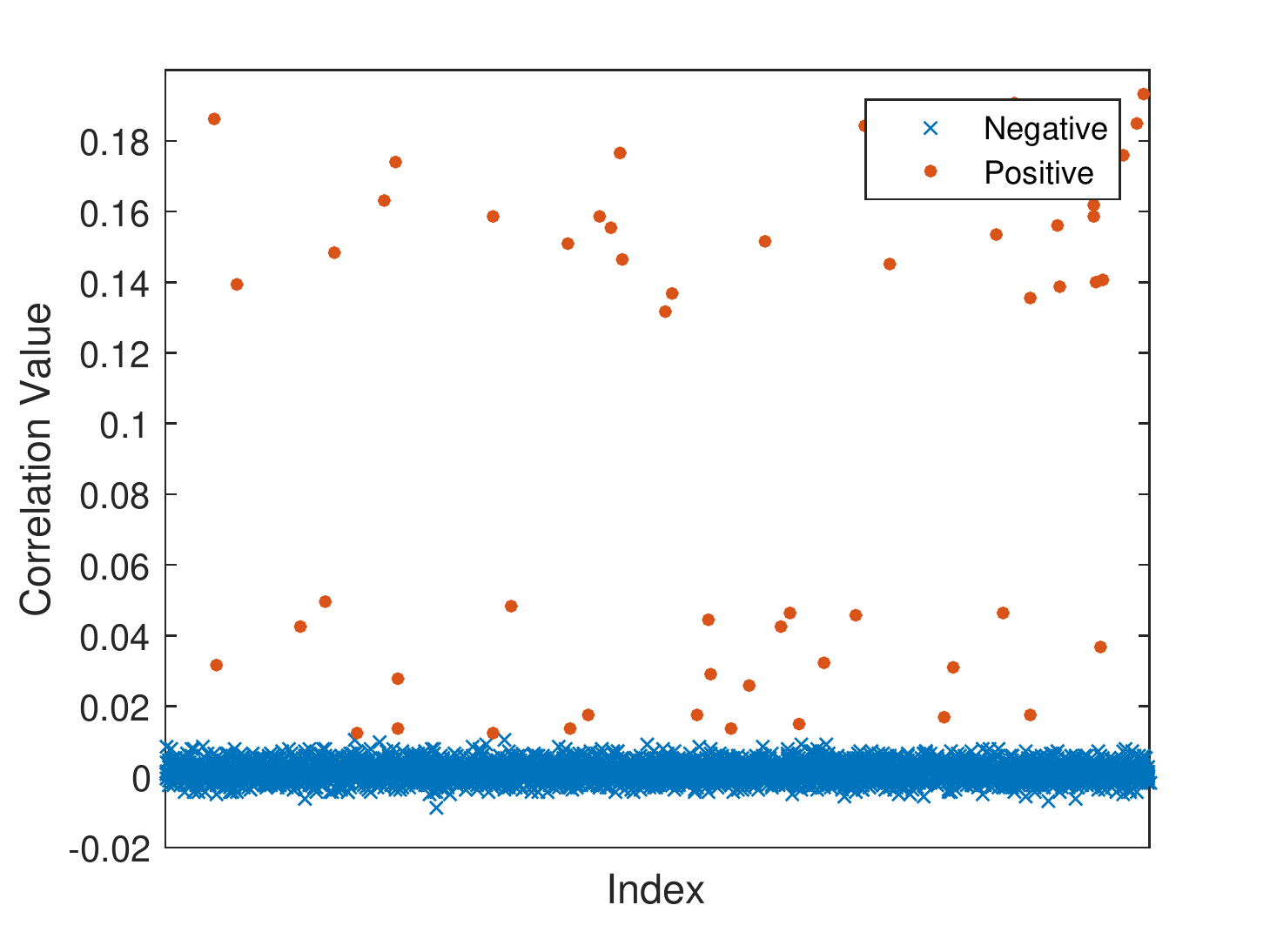}}
\subfloat[]
{\label{fig:subfig:ii}
\includegraphics[width=7cm]{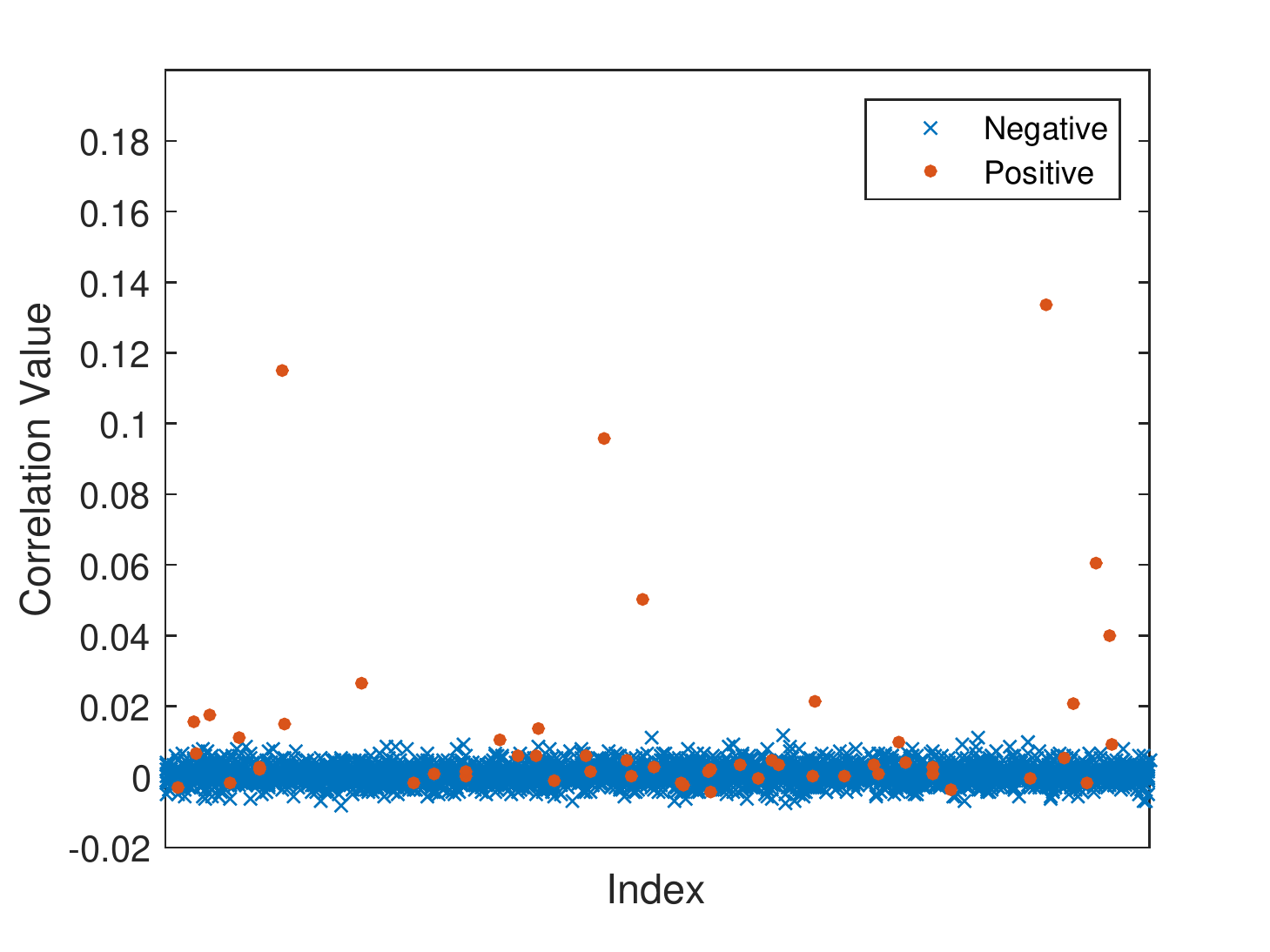}}\\
\subfloat[]
{\label{fig:subfig:iii}
\includegraphics[width=7cm]{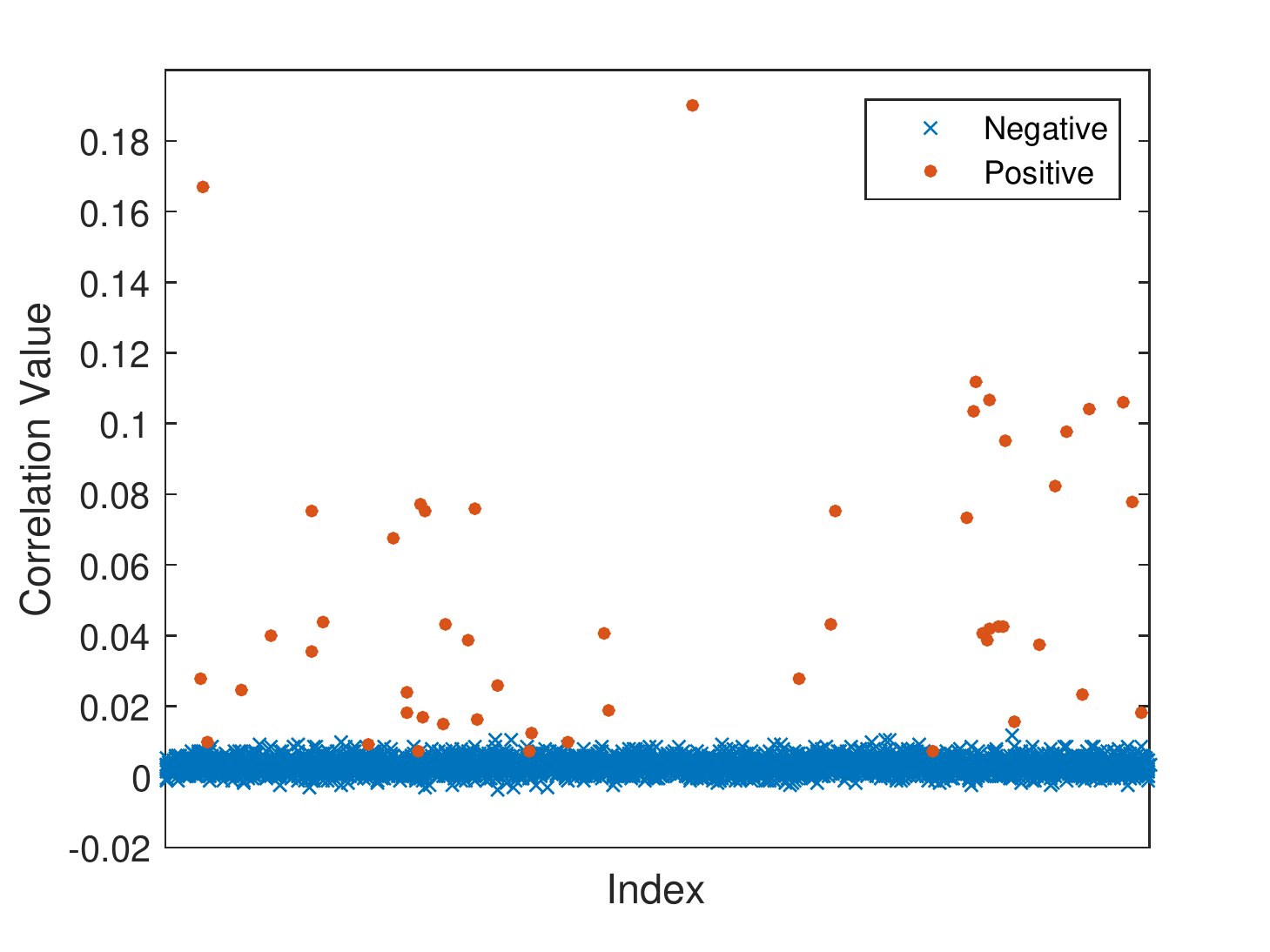}}
\subfloat[]
{\label{fig:subfig:iv}
\includegraphics[width=7cm]{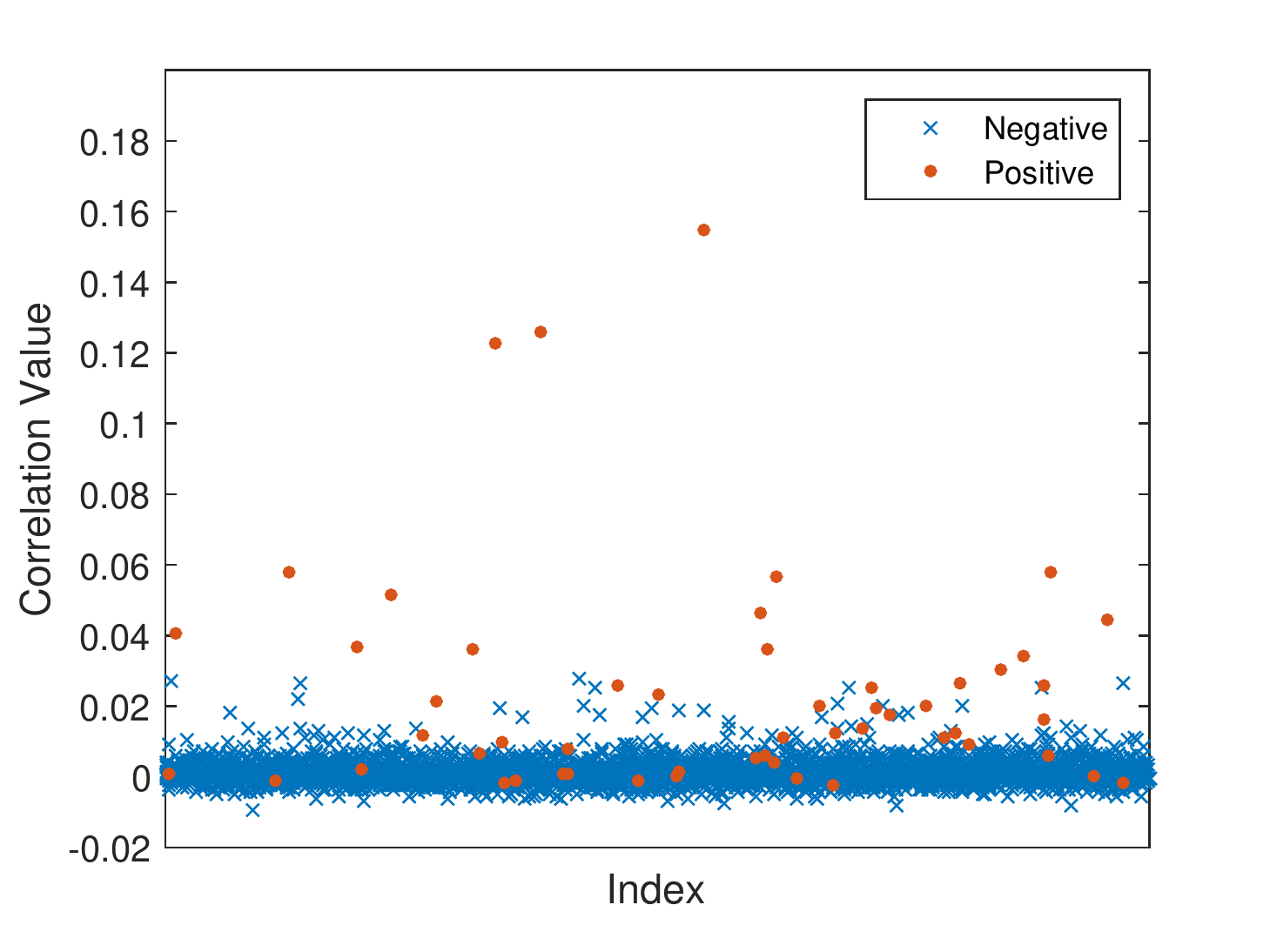}}
\caption{\small
Illustration of the feasibility of the new perspective and the key issues to be addressed. (a) single-camera case with single camera fingerprint; (b) two-camera case with single camera fingerprint; (c) two-camera case with two camera fingerprints; (d) two-camera case with two camera fingerprints and reposting confusion.
} 
\label{fig3}
\end{figure*}

Intuitively, we can individually extract camera fingerprints for all accounts by directly employing Eq.\ref{equ:3} and \ref{equ:4}. However, the online social networks are far more complex than what we think, and the users' behaviors are also varied. Therefore, we have to
take some practical issues into account when addressing the user identification problem.

The first is multiple camera problem. In real world, people generally own more than one photography device, \eg, cellphone camera, SLR camera. That is, the images in $\mathcal{I}_{i}$ may not be always captured by a unique camera. If we extract only one camera fingerprint to represent a user, it will cause some extra error. Therefore, it is necessary to model the user's camera feature by employing several camera fingerprints.

The second is reposting problem. Generally, some popular photos in an account are frequently reposted by other users. As shown in Fig.\ref{fig:reposting}, for example, users 1 and 2 don't belong to the same individual, but both of them repost the images from user 3. Without tackling this issue, users 1 and 2 will be incorrectly asserted to belong to the same individual, since reposted images have high probability to be captured by the same camera.

To clearly illustrate the feasibility of the new perspective and the key issues to be addressed, we conduct a series of experiments on a dataset with 1,576 images captured by 11 cameras. For each camera, all its images are randomly divided into two groups. Firstly, we verify whether camera fingerprints have the capability to reconcile the accounts belonging to the same individual. To this end, we simulate each group of images as the album of a user, and the groups (or users) deriving from the same camera are treated as positive pairs and others are negative pairs. After estimating the camera fingerprint for each simulated user, the correlation values between any two users are calculated by using  Eq.\ref{equ:4}.  The statistical results are illustrated in Fig.\ref{fig:subfig:i}, where red points indicate the correlation values between positive pairs and blue points are for negative pairs.  Clearly, positive pairs can be easily distinguished from negative ones. That is, camera fingerprint is indeed a kind of discriminative feature for distinguishing camera sources. However, an underlying assumption in this experiment is that the album in each account only contains images from a single camera. In more practicable scenarios, this assumption does not always hold, since people generally have more than one camera and frequently repost popular images from other cameras. Therefore, it is necessary to evaluate the effect of these confused factors. To construct a proper dataset, we combine any two groups from different cameras to simulate a user, and  two users who share one or two camera sources are treated as a positive pair and otherwise a negative pair. For each user, all the images from two cameras are used to estimate a unique camera fingerprint by using Eq.\ref{equ:3}.  The statistical results on correlation values are illustrated in Fig.\ref{fig:subfig:ii}, where many red points are mixed with blue points. That is, positive pairs cannot be clearly distinguished from negative ones. The reason lies in that using a unique  camera fingerprint to represent two cameras will unavoidably lead to confusion. Therefore, the discriminative capability of camera fingerprints are remarkably decreased. This is so-called multiple camera problem. If we can  identify camera sources of images in an account and individually estimate camera fingerprints for different camera sources, it is possible to avoid the multiple camera problem. To verify  this conclusion, we directly use the prior information of camera sources and estimate two camera fingerprints for each user.  To calculate the correlation value between any two users, we first calculate correlation values between any two camera fingerprints in two users, and select the maximum value as the final correlation value. The experimental results are shown in Fig.\ref{fig:subfig:iii}. As expected, the positive pairs are clearly distinguished from negative ones. That is, it is feasible to solve the multiple camera problem by estimating multiple camera fingerprints. However, introducing reposted images into album in an account will also lead to some confusion, Fig.\ref{fig:subfig:iv} illustrates this conclusion.

In brief, it is feasible to reconcile multiple users belonging to the same individual from the camera fingerprint perspective, and the key problems to be addressed are how to correctly extract multiple camera fingerprints and how to alleviate the effect of reposted images.

\begin{figure*}[t]
\centering
\includegraphics[width=16cm]{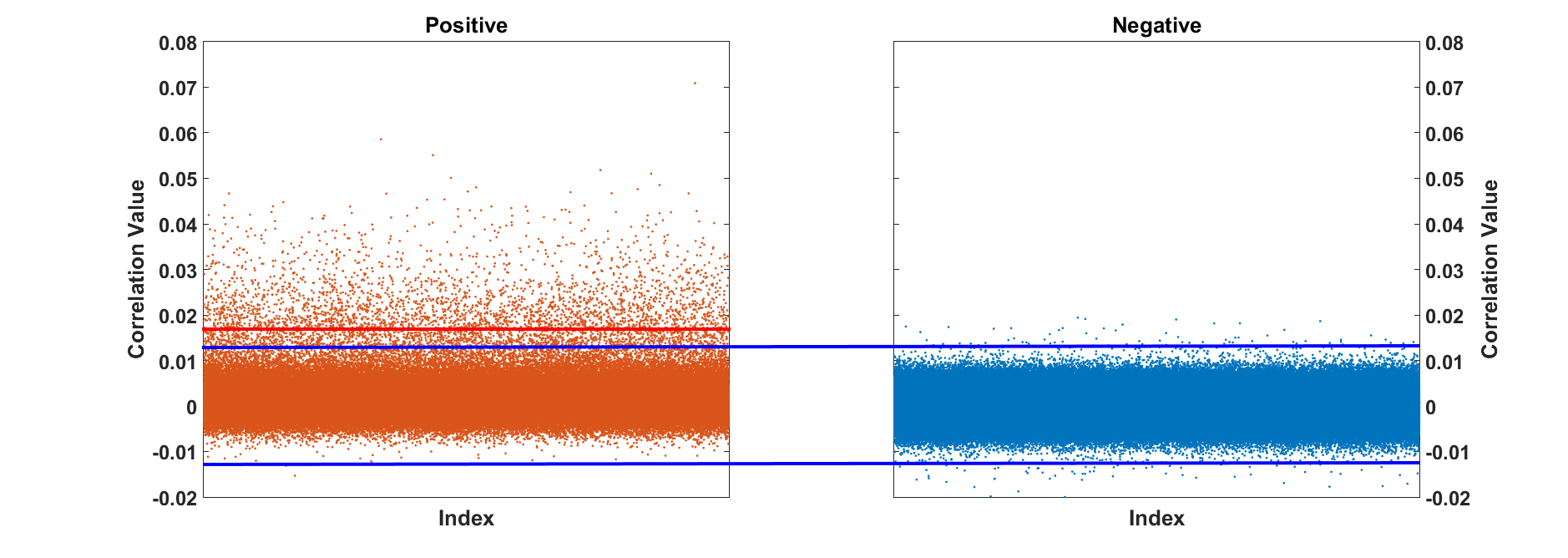}
\caption{\small
The correlation distribution of both positive and negative pairs. 
}
\label{fig:4}
\end{figure*}

\section{Methodology}
As discussed above, the key problem to be addressed is to overcome the confusion problem caused by the multiple cameras and reposted images. An intuitive solution is to first cluster images in an account into different groups according to their camera source information, and then individually estimate camera fingerprints for all groups. However, the camera source of each image in an account is unknown beforehand. Therefore,  we must design a method to avoid extracting multiple camera fingerprints in batch. In this section, we propose an incremental estimation approach of multiple camera fingerprints. 
The system framework is illustrated in Fig.\ref{fig:5}, which includes four key components: seed selection, incremental estimation, reposted image removing, and account matching.
\subsection{Seed Selection}
In essence, the proposed estimation approach is a tailored clustering method. Different from the scenarios of classic clustering methods, the multiple camera fingerprint estimation case requires accurate initial seeds (or initial camera fingerprints).
To obtain several accurate initial seeds, a pair correlation based method is proposed, which selects initial seeds by thresholding the pair correlation of residual noises of any two images.  

Specially, given an images set $\mathcal{I}_i=\{I_{i;1},...,I_{i;m}\}$, its noise pattern set $\mathcal{R}_{i}=\{R_{i;1},...,R_{i;m}\}$ can be obtained correspondingly by Eq.(\ref{equ:3}), and the correlation values between any two images' residual noises are calculated by Eq. (\ref{equ:4}). For any correlation value that is greater than a predefined threshold, the corresponding two images are grouped together, and a camera fingerprint is estimated as a seed for further steps. In fact, the seed selecting method is based on the following assumption:
\\\\
\textbf{Assumption 1}: 
If the correlation of two images' residual noises is high enough, the two images have high probability to be captured by the same camera. 
\\

To verify this assumption, a database containing 1,576 images captured by 11 cameras is used. For any image pair from the database, if two images are captured by the same camera, we call it a positive pair, otherwise a negative pair. The correlation value of two images in any pair is separately calculated by using their residual noises, and their results are illustrated in Fig.\ref{fig:4}. 
As we can see, the correlation distribution of negative pairs are limited to a low and narrow range, while positive pairs are scattered in a higher but larger range. In addition, there is a narrow but dense overlap for negative and positive image pairs, as indicated by two blue dot lines. 

In a sense, it seems to be conflict with the conclusion in Fig.\ref{fig:subfig:i}. In fact, the positive or negative pairs in this case are quite different from these in Fig.\ref{fig3}. In Fig.\ref{fig3}, a pair is consist of two groups of images from two cameras, and a camera fingerprint is estimated from each group by using Eq.\ref{equ:3}. Since the noise component ,which is Gaussian white noise in essence \cite{Chen2008}, can be smoothed during the averaging procedure, stable and reliable camera fingerprints can be achieved. In contrast, a pair contains only two images here. In this way, the camera fingerprints here are essentially the residual noises of the images and
Gaussian white noises are not removed. Therefore, the correlation between two images in a pair cannot be estimated correctly. That is, employing only residual noise cannot completely distinguish images coming from different cameras. Fortunately, we can also observe that a pair of images has a high probability to be captured by the same camera if their pair correlation value is high enough, as indicated by the points above the red dot line in Fig.\ref{fig:4}. These observations support our assumption.

In brief, we can choose some stable and reliable seeds from images in an account by using the residual noises and setting a high threshold value. It should be noticed that setting high threshold value will lead to high false rejection problem, i.e., many positive image pairs are treated as negative ones. We will leave this problem to the next sub--section.

{
\subsection{Incremental Estimation of Multiple Camera Fingerprints}

Our goal is to put all images captured by the same camera into a group so as to estimate a reliable camera fingerprint. To do that, we propose an incremental camera fingerprint estimation method, which includes three key steps: initializing camera fingerprints, merging consistent groups, and estimating new camera fingerprints. 

\noindent\textbf{Initializing camera fingerprints}

Using the above-mentioned seed selecting method, a number of positive image pairs are chosen and each image pair is separately treated as a group. As shown in the step 1 in Fig.\ref{fig:5}, four positive image pairs are selected as seeds. That is, the user is initially assumed to have four different cameras and each camera has only captured two images in corresponding group. More formally, let  $\mathcal{C}_i$ denote the set of all different groups (or Clusters) in $i^{th}$ user and $C_{i;j}$ represents $j^{th}$ group. Then, the set of camera fingerprints $\{ S_{i;j} \}$ can be estimated individually from these groups by Eq.\ref{eq.6},

\begin{equation}
\label{eq.6}
S_{i,j}= \frac{\sum I_{i;l}\cdot R_{i;l}}{\sum I_{i;l}^2},\;\;\; l:I_{i;l}\in C_{i,j},
\end{equation}
Here, we denote  $\mathcal{S}_{i}$ as the set of camera fingerprints  $\{ S_{i;j} \}$. 

Initially, all positive image pairs are treated individually as groups, and one camera fingerprint is estimated from each group. In this way, total $|C_i|$ camera fingerprints are obtained, which are used as initial seeds. 

\noindent\textbf{Merging consistent groups}

As discussed above, all positive pairs are selected by thresholding the correlation value of residual noises of two images. Although the two images belonging to the same pair have high probability to be captured by the same camera, we cannot ensure that any two pairs of images are captured by totally different two cameras. In other words, some image pairs may share the same camera source. Therefore, it is necessary to merge these groups. Toward this end, a similar strategy to seed selection is employed here to merge consistent groups. In particular, given any two groups  $C_{i;j},C_{i;k} (j<k)$, we merge them into one cluster if correlation value of their corresponding camera fingerprints $S_{i,j}$, $S_{i,k}$ is greater than a pre-defined threshold $\alpha$. Formally, the updating procedure can be formulated as

\begin{equation}
\label{equ:10}
C_{i;j} = C_{i;j} \bigcup C_{i;k},  \ if \ corr(S_{i;j},S_{i;k}) > \alpha,
\end{equation}
where $C_{i,j}$ is the merged image set. That is, any image pair which correlation is greater than $\alpha$ will be merged into the same group.

After group merging, a new set $\mathcal{C}_i$ of groups is generated, and an updated camera fingerprint set $\mathcal{S}_{i}$ can be estimated. In fact, group merging will lead to two benefits. 
First, some redundant camera fingerprints are removed, which will result in a more compact set of feature patterns. Second, a more reliable camera fingerprint for a unique camera can be estimated, since more samples are collected individually for cameras and used to smooth Gaussian white noise. Therefore, using group merging procedure will lead to more reliable user identification. 

\noindent\textbf{Estimating new camera fingerprints}

In order to obtain reliable and stable seeds, $\alpha$ is generally set to a high value. In this way, many images that are captured by one of cameras associated with $C_{i;j}$ will be rejected, while the selected image pairs are guaranteed to be true positives. It is useful for generating accurate camera fingerprints to correctly reassign these rejected images into correct groups. An intuitive and straightforward solution is to assign 
each rejected image to the most similar group by estimating the correlation values between the residual noise of rejected image and all groups' camera fingerprints. However, if some rejected images are incorrectly assigned to one group, it will result in some false acceptation. In this case, images from different cameras will be in the same group, which leads to unreliable camera fingerprints. To avoid incorrect acceptation, we further pre-define a threshold  $\beta$ to determine whether we assign a rejected image into a group. Formally, given any rejected image's residual noise $R_{i;k}$, we assign the image $I_{i;k}$ into the group $C_{i;j}$ if the maximum correlation value achieved between $R_{i;k}$ and $S_{i,j}$ is greater than $\beta$. 

Once the reject image $I_{i;k}$ is assigned to the group $C_{i;j}$,  $S_{i,j}$ will be updated by using the new image set $C_{i;j}$. Repeating these steps will assign most of rejected images to corresponding groups, and more accurate camera fingerprints will be estimated incrementally for these groups.

\begin{figure*}[t]
\centering
\includegraphics[width=18cm]{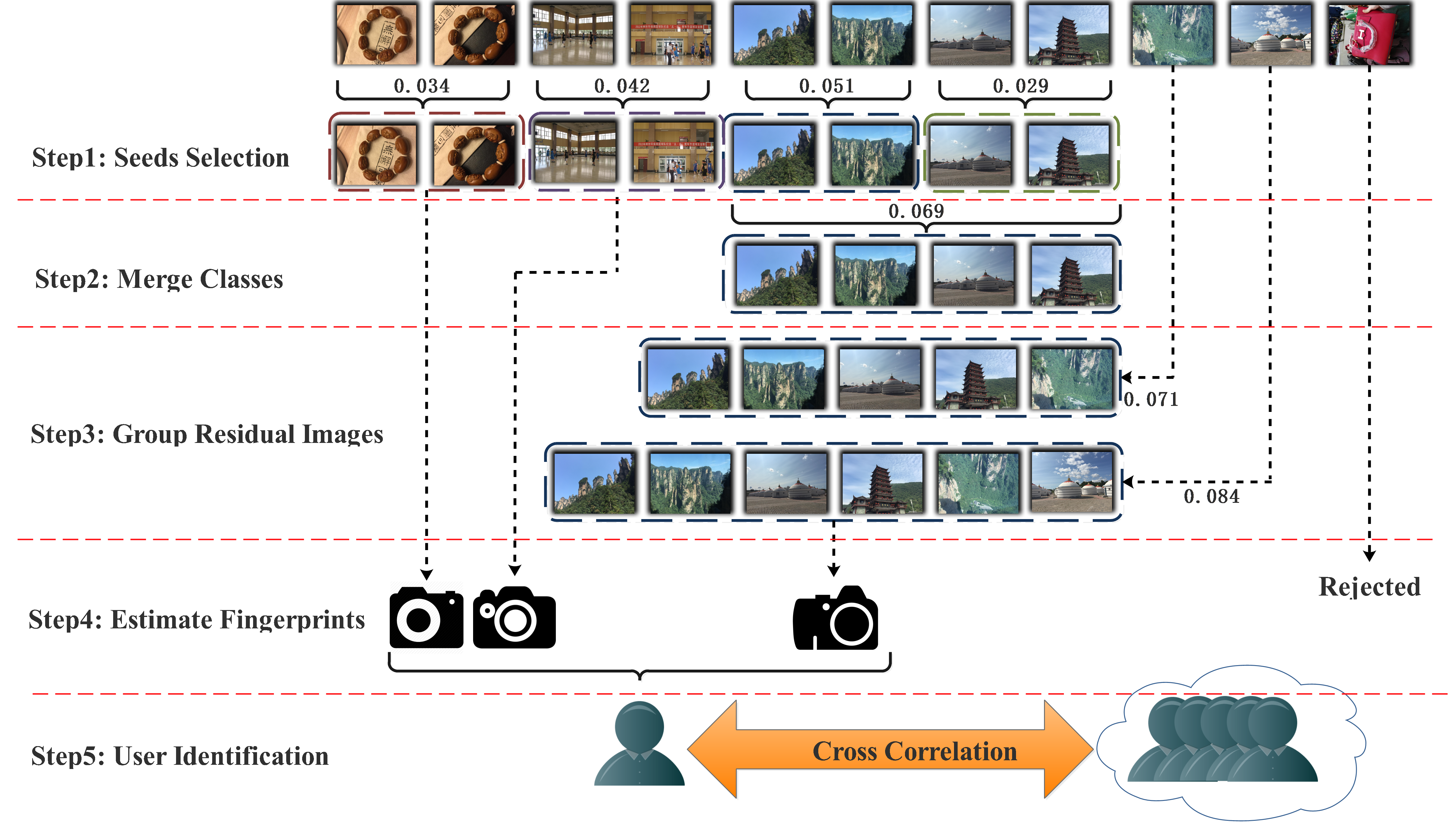}
\caption{\small
The framework overview of our approach (UCI).
} 
\label{fig:5}
\end{figure*}

\noindent\textbf{Complete Algorithm}

The complete algorithm of multiple camera feature estimation is listed in Algorithm \ref{alg1}. Generally, there are an initialization step and an iterative step. In the first step, possible positive image pairs are selected to individually form a set of groups $\mathcal{C}_i =\{ C_{i,j}\}$, and the corresponding camera fingerprints $\mathcal{S}_i =\{ S_{i,j}\}$ are extracted as seeds. Then, both group merging and rejected image reassigning steps are iteratively performed to incrementally improve the reliability of camera fingerprints.

\begin{algorithm} 
\caption{User Camera Fingerprint Estimation} 
\label{alg1} 
\begin{algorithmic}[1]
\REQUIRE Given an images set $\mathcal{I}_{i}$ of user $i$, thresholds $\alpha, \beta$;
\STATE Get residual noise set $\mathcal{R}_i$ by Eq. \ref{eq.2};
\STATE Calculate $corr(R_{i,j},R_{i,k})$, where $R_{i,j},R_{i,k}\in\mathcal{R}_i$;
\STATE Initializing the number of camera fingerprints $J=0$, \\
group set $\mathcal{C}_i=\{\}$ and camera fingerprint set $\mathcal{S}_i=\{\}$;
\FOR{\textbf{all} $corr(R_{i,j},R_{i,k})$}
\IF{$corr(R_{i,j},R_{i,k})\geq \alpha$}
\STATE $J\leftarrow J+1$;
\STATE $\;\mathcal{C}_{i}\leftarrow C_{i,J} = \{I_{i;j},I_{i;k}\}$;
\STATE $\mathcal{I}_i \leftarrow \mathcal{I}_i-\{I_{i;j},I_{i;k}\}$;
\STATE $\mathcal{S}_{i} \leftarrow S_{i,J}$, where $S_{i,J}$ is estimated by Eq. \ref{eq.6};
\ENDIF
\ENDFOR
\STATE Set $\Delta=1$;
\WHILE{$\Delta$}
\STATE Merge consistent groups $\mathcal{C}_i$ by Eq. \ref{equ:10};
\STATE Calculate $\rho=\max(corr(R_{i;j},S_{i;k}))$,\\
 where $j:I_{i;j}\in\mathcal{I}_i$, $k:C_{i;k}\in\mathcal{C}_i$;
\IF{$\rho\geq\beta$ \textbf{and} $\mathcal{I}_i\neq\varnothing$}
\STATE $C_{i;k}\leftarrow C_{i;k}\bigcup\{I_{I;j}\}$;
\STATE $\mathcal{I}_i = \mathcal{I}_i-\{I_{i;j}\}$;
\STATE Update $S_{i;k}$ by Eq. \ref{eq.6};
\ELSE 
\STATE $\Delta=0;$
\ENDIF
\ENDWHILE
\ENSURE Camera fingerprints set $\mathcal{S}_i$
\end{algorithmic} 
\end{algorithm}

\subsection{Dealing with Reposted Images}
As we mentioned before, reposted images may cause remarkable confusion. Therefore, it is necessary to take reposted images into account when extracting multiple feature patterns to represent a user. However, reposting behaviors are quite complex and it is difficult to distinguish them from normal images. Therefore, we attempt to address the problem by investigating the reposting behaviors. 

However, the reposting behaviors are both complex and uncertain due to the diversity of users. In order to simplify the difficulty of modeling, we consider only two common reposting behaviors. First,  users repost many images, and these images come from different sources (e.g., from several different accounts). In this case, the number of reposted images from a specific camera source are far less than user's own images. Second, users repost multiple images from a single source (e.g., an account). In this case, the reposted images from a specific camera source are still scarce, since the reposted images are more likely be captured by different cameras of original user. In brief, we assume that the reposted images are from different cameras and the number of reposted images captured by the same camera are scarce. Based on this assumption, we can efficiently suppress the confusion of reposted images.

In fact, the algorithm of multiple camera feature estimation discussed above has already had a certain capability of eliminating reposted images, since it can directly reject the reposted images who don't meet the condition of positive image pair. For example, when people repost only a single image from other user's album, it can neither be chosen as a seed nor be assigned into a group by the proposed algorithm. Therefore, this kind of reposted image has no effect on the estimation of users' camera features. However, when more than one reposted image is captured by the same camera, it will lead to some mistakes. For these reposted images, they may be chosen as seeds at the initial step, and most of them are grouped into the same group after several iterations.
To address this issue, we adopt a simple but effective post-processing step to alleviate their effect.
According to the observation on users' reposting behaviors, the number of reposted images are generally scarce for a specific camera. Therefore, the camera fingerprint set $\mathcal{S}_{i}$ achieved by incremental estimation steps can be refined as follows

\begin{equation}
\label{eq:11}
\mathcal{S}_{i}=\{S_{i;j}\;|\;j:|C_{i;j}|\geq \lambda\}.
\end{equation}
where $|C_{i;j}|$ means the number of images in class $C_{i;j}$, and $\mathcal{S}_{i}$ can be taken as the set of camera fingerprints of user $i$.

For any pairwise of social users, we assert them to belong to a single individual if their feature correlation is greater than a predefined threshold. In this way, the groups with few images are treated as reposted image sets and filtered out.

\subsection{Similarity Estimation}
In this section, we mainly discuss about how to identify multiple accounts belonging to the same individual by estimating their similarity based on their camera features. 
Using multiple camera fingerprint estimation algorithm, a  camera feature set $\mathcal{S}_i$ is obtained, and element $S_{i;j}\in\mathcal{S}_i$ denotes one camera fingerprint.
As we mentioned above, we consider two users belong to a single individual if they share at least one camera. Therefore, the problem is changed to determine whether two users have similar members of their camera features. Formally, the problem can be represented as follows. For any two users $U_x$ and $U_y$, their camera features $\mathcal{S}_x=\{S_{x;1},...,S_{x;m}\}$ and $ \mathcal{S}_y=\{S_{y;1},...,S_{y;n}\}$ are obtained.
We estimate their similarity by using the maximum correlation value between two camera fingerprints in  $\mathcal{S}_x$ and $\mathcal{S}_y$, which can formulated as follows:

\begin{equation}
\textit{\textbf{d}}(U_x, U_y)=\max\{\;\mathop{corr} \limits_{i,j} (S_{x;i},S_{y;j})\;\},
\end{equation} 
where $\textit{\textbf{d}}(U_x, U_y)$ denotes the similarity between $U_x$ and $U_y$. 

\section{UID-BJTU: A Dataset for user identification based on camera fingerprint}
To the best of our knowledge, it is the first  time to re-inspect the user identification problem from the perspective of camera fingerprint. Therefore, no public dataset is available for testing.
In order to evaluate the performance of proposed scheme, we 
collect and construct a new benchmark, named UID-BJTU.
This benchmark is consist of two different collections, \ie, simulated user dataset and online social user dataset. The images in former dataset are acquired from multiple cameras directly, which have clear information of camera sources. This dataset is mainly used to clearly evaluate the effectiveness of each stage in our approach. Instead, the images in the latter one are crawled from online social networks, which come from real online users. Details about the benchmark is described in this section.

\subsection{Simulated User Datasets}
In order to comprehensively evaluate effectiveness of the proposed approach, we need know the accurate information about users, such as the relationship among users, the number of cameras, the number of images captured by any camera. 
However, it is not an easy task to obtain an ideal evaluation benchmark by crawling the data of online users. Generally speaking, an image's camera source can be determined by the information provided by Exchangeable image file format (Exif), which contains camera series number, brand and model information.
Unfortunately, most of these images' Exif files are incompletely due to various reasons such as privacy, post-processing, so it is impossible to construct a reliable camera source groundtruth from online user dataset.
Therefore, in order to evaluate the performance of proposed incremental estimation scheme, 
we directly acquire a set of images by using multiple pre-selected cameras to
simulate the data of online users.
Specially, total 11 cameras are collected and 1,576 images with clear camera source information are acquired.

To guarantee data diversity and avoid potential confusion of camera brand and model, we take camera brand and model into account when choosing cameras. Table \ref{table:1} lists the details of cameras' brands and the number of images from each camera.

\begin{table}[h]
\centering  
\caption{Summary of Camera Sources}
\vspace{10pt}
\renewcommand\arraystretch{1.5}
\begin{tabular}{cccc}  
\hline
\hline
NIKON D7000 & DVTs & HUAWEI & iPhone6 Plus\\
125 & 117 & 200 & 106\\
\hline
NIKON D7000 & iPhone6 & Canon 900Ti & Canon 650D \\
112 & 250 & 146 & 109 \\
\hline
HM-NOTE & NIKON D7000 & PENTAX K-50& -\\
139 & 278 & 266& -\\
\hline
\hline \label{table:1}
\end{tabular}
\end{table}

The key advantage of the proposed method is to significantly alleviate the confusion problem from multiple cameras and reposted images. Therefore, in order to clearly evaluate the effectiveness of the proposed approach, we need to simulate users with diverse camera sources and reposted images. In our experiments, three kinds of users are simulated, which are as follows:

\noindent\textbf{Offline}$_1$ \textbf{Dataset}: One of 11 cameras corresponds to a single individual, and all images from the camera is randomly divided into two groups. Each group of images simulates one user of the individual. If two users are derived from the same camera, they are treated as a positive pair, otherwise a negative pair. In this way, total 22 users, 11 positive pairs and 220 negative pairs are constructed. In order to involve the reposed images, total 110 images from an extra camera are randomly added into 22 users.

\noindent\textbf{Offline}$_2$  \textbf{Dataset}:  Any combination of two cameras corresponds to a single individual, and all images from the two cameras are randomly divided into two groups.  Each group of images simulates one user of the individual. The two users belonging to the same individual is called a positive pair. For any two users, if they don't share any camera source, we call them a negative pair.  In this way, total 110 users, 55 positive pairs and 3,960 negative pairs are constructed. Similarly,  in order to involve the reposed images, total 550 images from an extra camera are randomly added into 110 users.

\noindent\textbf{Offline}$_3$  \textbf{Dataset}:  The constructing process is quite similar with Offline$_2$. The main difference lies in that a combination of three cameras corresponds to a single individual and total 550 images from an extra camera are randomly added into 110 users.

A summary about these datasets is listed in Table \ref{table:2}.



\begin{table}[h]
\centering  
\caption{Statistics on All Datasets}
\vspace{10pt}
\renewcommand\arraystretch{1.2}
\begin{tabular}{lcccc}  
\hline
 & \textbf{Offline}$_1$ & \textbf{Offline}$_2$ & \textbf{Offline}$_3$ &  \textbf{Online} \\ 
\hline 
\small{Camera Source} & 1 & 2 & 3 & - \\
\small{Generated user} & 22 & 110 & 110 &  192 \\
\small{Positive pairs} & 11 & 55 & 55 &   96\\
\small{Negative pairs} & 2,20 & 3,960 & 2,060 &  18,240 \\
\small{Sum of Reposted} & 110 & 550 & 550&  960 \\
\hline \label{table:2}
\end{tabular}
\end{table}

\subsection{Online Social User Dataset}
The final goal of proposed identification method is to identify online user's identity by employing camera fingerprints, so it is necessary to evaluate the performance on real online data. However, it is difficult to construct a reliable groundtruth for online users' identity. Therefore, we attempt to construct a simulated network with controllable data. 

Toward this end, we first crawled 15,328 images from 96 Flickr users' albums, and the image number of users are varied from 82 to 250. Although the simulated network with 96 users is far from a large-scale dataset, we can manually ensure that any pair of these users don't belong to the same individual and no any reposted images are included in these albums. That is, one original user corresponds to a single individual. To construct positive pair of users belonging to a single individual, we randomly divide the images in its original user into two groups, where each group is treated as a new user. In this way, each individual corresponds to two users,  which are treated as a positive pair. Totally, 192 users, 96 positive pairs and 18,240 negative pairs are constructed. In addition, 107 images from an independent user are treated as reposted images and randomly added to 192 users.

\section{Experimental Evaluation}
In this section, we conduct several experiments to evaluate the proposed user identifying methods. In particular, the grouping performance of the proposed method on \textbf{offline} datasets are first evaluated, and then the reposted image problem is verified. Finally, we test our algorithm on \textbf{online} benchmark and give some useful conclusions. It is worth noting that the perspective of the proposed framework is totally different from the traditional user identification frameworks. Therefore, we cannot compare the proposed approach with previous works due to the lack of testing database containing both camera fingerprint and user public profiles or activity characters. 

\subsection{Metrics and Baseline}
Before we present our experimental results, we first introduce several metrics to evaluate the algorithm's performance. 
In the proposed approach, the incremental grouping method plays very important role. Therefore, in addition to evaluate the  final identification, we should also fully evaluate the grouping effectiveness.
The grouping is the foundation of the proposed method, which aims to identify the camera sources of images. Although the proposed grouping algorithm is quite different from classical clustering algorithm, we can still employ the evaluation metrics of classical clustering methods. Here, we employ purity, precision and recall to evaluate the grouping performance. 

Purity is a simple and transparent measure for evaluating the performance of clustering. Given the incremental estimation grouping result $\mathcal{C}=\{C_{i,1},C_{i,2},...,C_{i,J}\}$ and the groundtruth of camera source $\mathcal{C}'=\{C'_{i,1},C'_{i,2},...,C'_{i,K}\}$, purity is defined as follows

$$purity(\mathcal{C},\mathcal{C'})=\frac{1}{N}\sum_{j} \max_k|C_{i,j}\cap C'_{i,k}|,$$
where $N$ denotes the number of images in user $i$. It should be noticed that high purity is easy to achieve when the number of clusters is large enough (for example, purity is to 1 if each image is grouped a single group). To address this problem, we employ precision and recall as the additional measures to further evaluate the performance. Formally, 

$$\textit{Precision}=\frac{TP}{TP+FP},\,\,\,\,\,\,\textit{Recall}=\frac{TP}{TP+FN},$$
where $TP$ (true positive) decision assigns two images from the same camera to the same group, $FN$ (false negative) decision assigns two images from the same camera to different groups, and $FP$ (false positive) decision assigns two images from different cameras to same group. More details can be found in \cite{Introduction2008}. Briefly, a high precision means that images are mostly assigned to the true group, and high recall reflects that most images are grouped. In our scenario, we mainly pursuit a high precision so as to guarantee that estimated camera fingerprint is not influenced by multiple camera sources. After that, we allow recall to be a little lower, since not every image is necessarily grouped.

To evaluate the user identification performance, we employ true positive rate and false positive rate. Furthermore, we also take user identification problem as a user retrieval problem, and employ Mean Average Precision (MAP) to evaluate the identification performance. 

To fully show the feasibility of the proposed framework, we design three schemes for comparisons.
\begin{itemize}

\item{\textbf{SCF} (\textbf{S}ingle \textbf{C}amera \textbf{F}ingerprint)}: It assumes that all images  in a user's album are captured by the same camera. Under this assumption, each user is represented by only one camera fingerprint estimated from all images. By estimating the similarity among different users, we can determine whether two users belong to the same individual or not.
Compared with the proposed approach, this method take neither the multiple camera problem nor reposting behaviors into account. 

\item{\textbf{MCF} (\textbf{M}ultiple \textbf{C}amera \textbf{F}ingerprints):} This scheme incrementally divides images of a user into several groups, and individually estimates camera fingerprints for groups by using Eq. \ref{eq.6}. Then, the groups that meet the threshold $\gamma$ are removed, and all the other groups' camera fingerprints are treated as the user's camera feature. \textbf{MCF} takes only multiple camera problem into account.

\item{\textbf{UCI} (\textbf{U}ser \textbf{C}amera \textbf{I}dentification)}: In this scheme, both multi-camera problem and reposting behaviors are taken into account.
\end{itemize}

\subsection{Grouping Performance Evaluation}

\begin{table}[h]
\centering  
\caption{Clustering Performance Evaluation.}
\vspace{10pt}
\renewcommand\arraystretch{1.5}
\begin{tabular}{lccc} 
\hline
& \textbf{Offline}$_1$ & \textbf{Offline}$_2$ & \textbf{Offline}$_3$  \\ 

\hline
Purity  & 0.76 (0.91) & 0.54 (0.76) & 0.45 (0.48)  \\         
Precision  & 0.91 (0.83) & 0.90 (0.62) & 0.85 (0.35)  \\      
Recall  & 0.72 (1.00) & 0.50 (1.00) & 0.42 (1.00)  \\
\hline \label{table:4}
\end{tabular}

\end{table}

The purpose of grouping is to accurately divide all images in a user into different groups according to their camera sources. In this way, a camera fingerprint can be estimated for each camera source and multiple camera problem can be alleviated significantly. In fact, high precision of clustering is more important than purity and recall in our case, since high precision means that there are less confusing images in a group belonging to a single camera source. In this section, we conduct a series of experiments to verify the effectiveness of the proposed incremental clustering method, and the experimental results are listed in Table \ref{table:4}. To clearly show the advantage of the proposed method, we also provide a baseline. In this baseline, we assume that all images of one user belong to the same camera source. That is, all images of one user is treated as a unique group. The statistical results of baseline on purity, precision and recall are associated with the results of the proposed method in brackets for comparisons. Clearly, for the \textbf{Offline}$_1$ dataset, all the indicators of baseline outperform the proposed method, since each user in this dataset has only a camera source. When more cameras are introduced into users, the recall of the baseline keeps perfect one, which obviously better than the proposed method. 
We also notice that the purity of the proposed method is also worse than the baseline. However, it does not prove that the performance of the baseline is better. As we mentioned before, not all images are assigned to one of groups during the incremental clustering procedure, since some images whose correlations to any group centroid are lower than iteration terminal criterion $\beta$ are filtered out. Therefore, the recall and purity will be inevitably decreased when we calculate them by taking all image into account. If the rejected images are not counted in, high recall and precision can be obtained. For example, when we remove these rejected images from users, and the recalculated recall values are 0.9524, 0.9136, 0.9197, respectively. In our scenario, however, how to alleviate the mutual confusion among multiple cameras is more important than providing high recall. Therefore, the key of a good grouping method is to ensure that the images in a group are accurately derived from the same camera. 
\begin{figure*}[t]
\centering
\subfloat[]
{\label{fig:subfig:1}
\includegraphics[width=7cm]{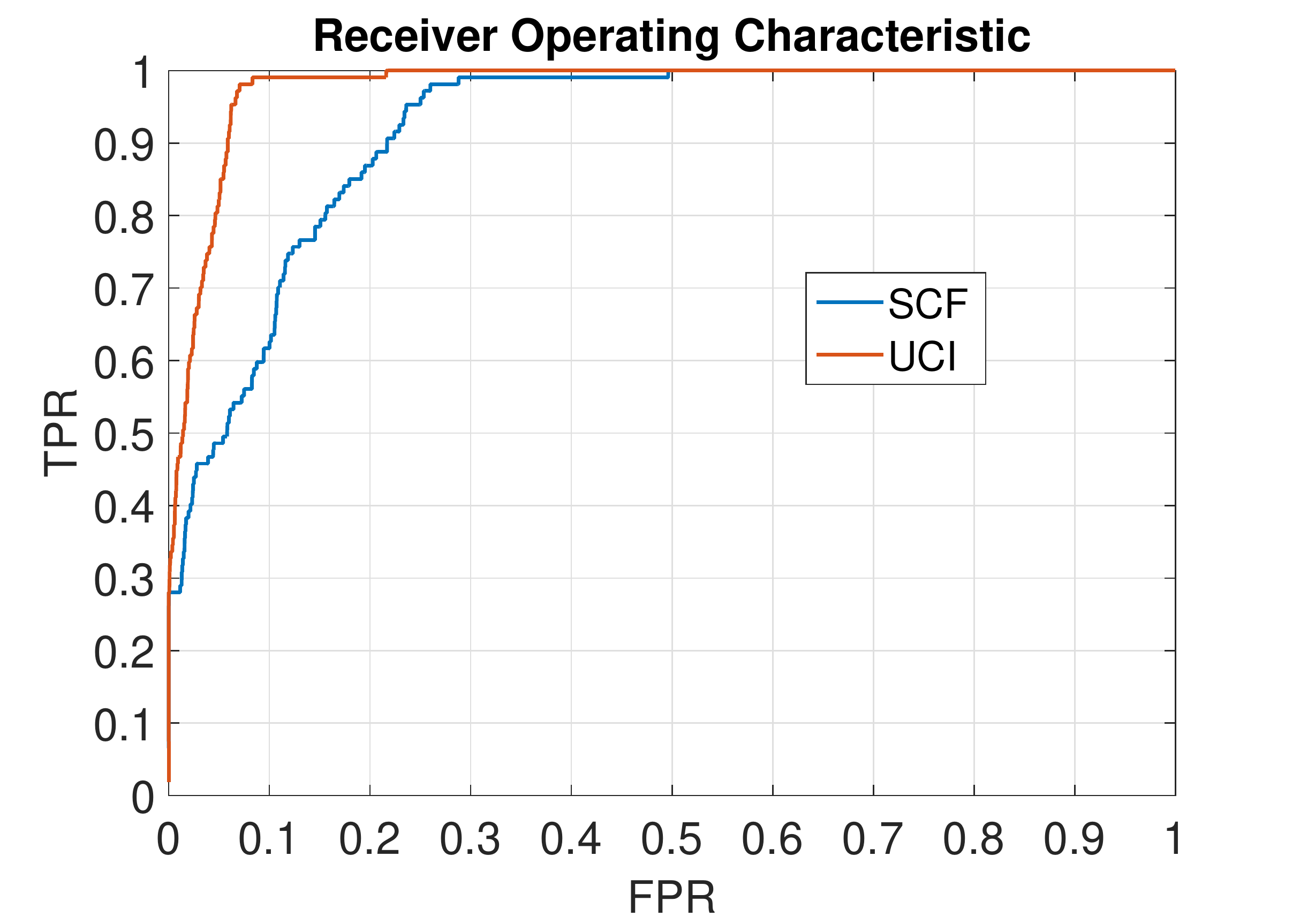}}
\subfloat[]
{\label{fig:subfig:2}
\includegraphics[width=7cm]{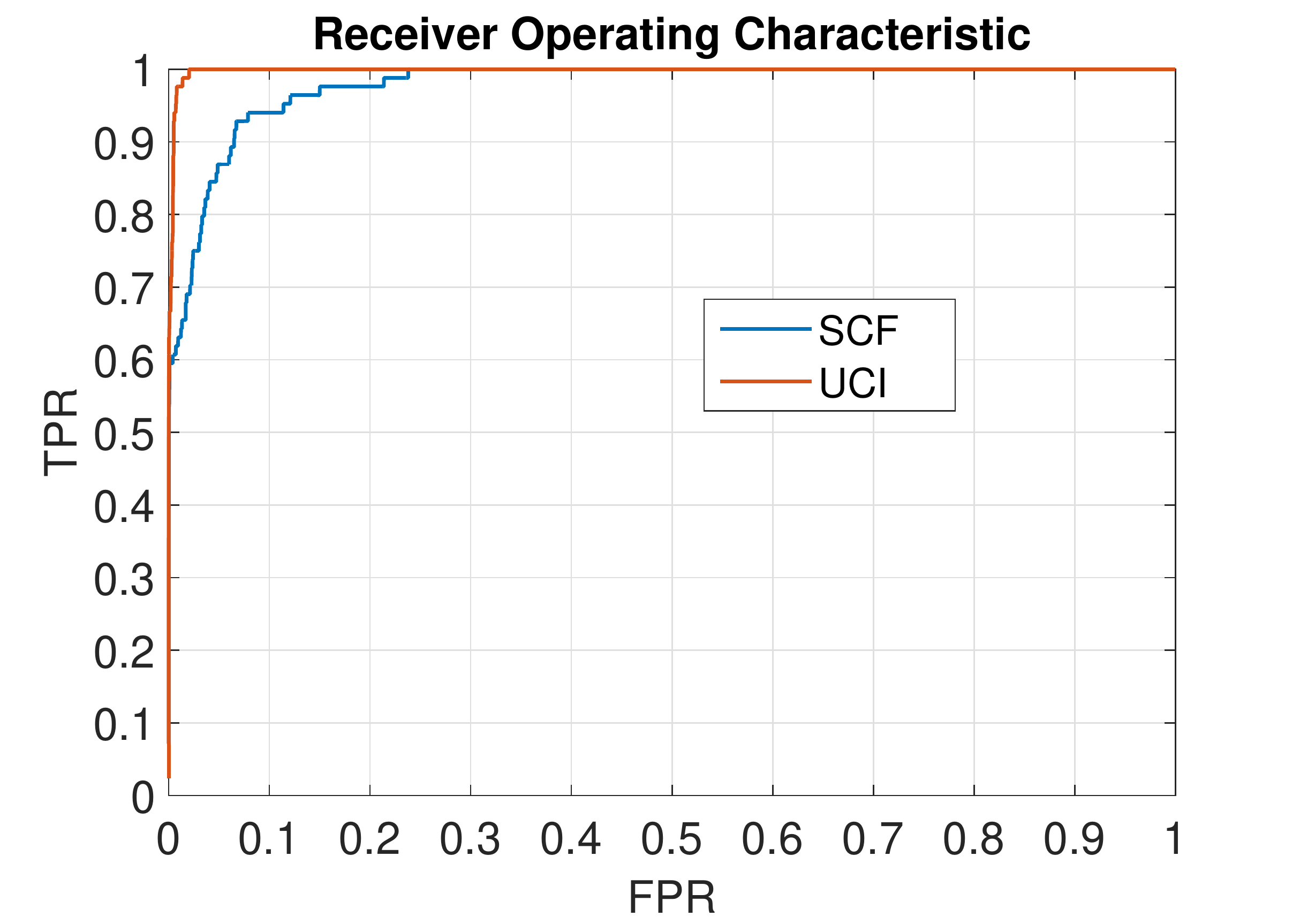}}\quad
\subfloat[]
{\label{fig:subfig:4}
\includegraphics[width=7cm]{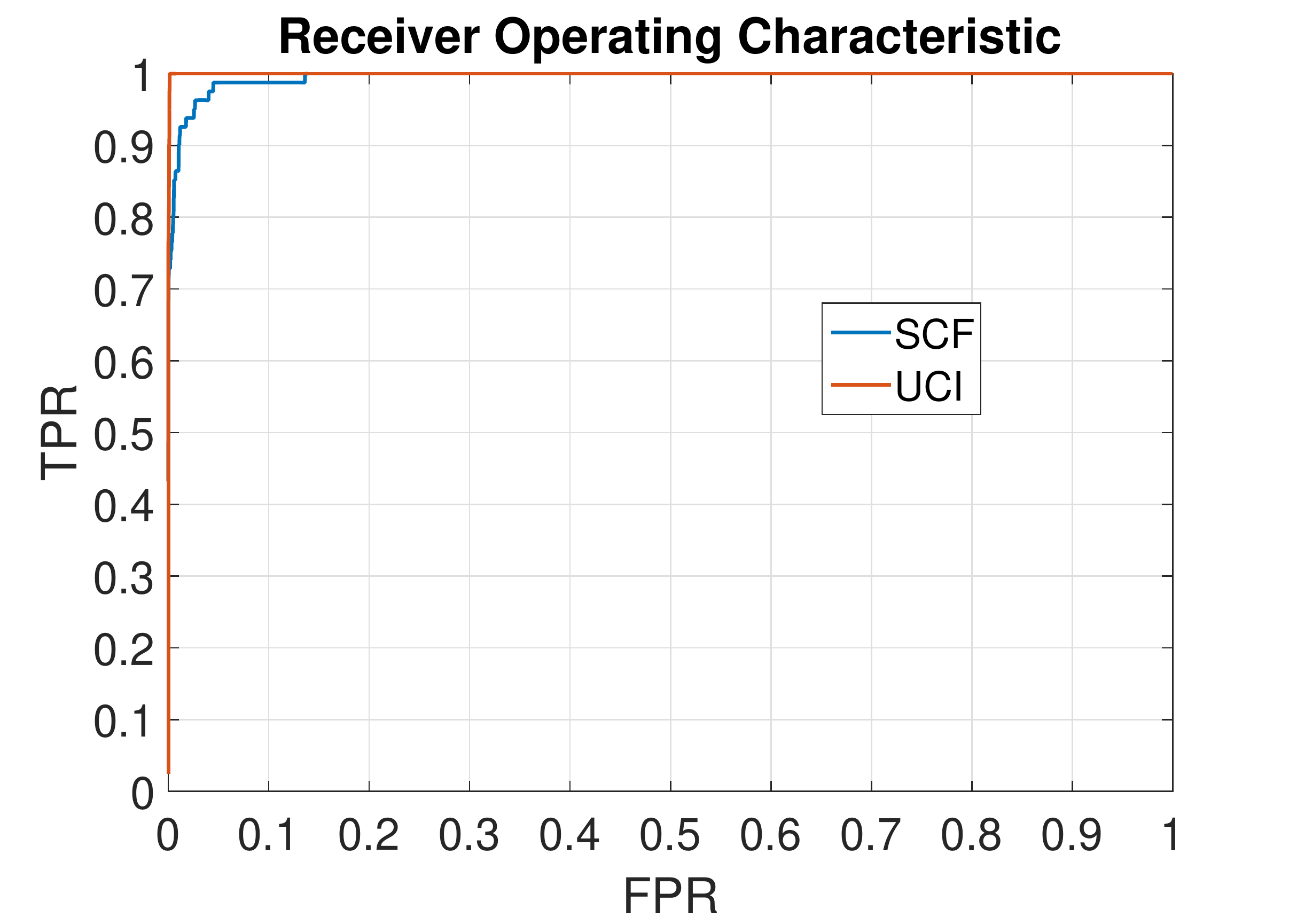}}
\subfloat[]
{\label{fig:subfig:5}
\includegraphics[width=7cm]{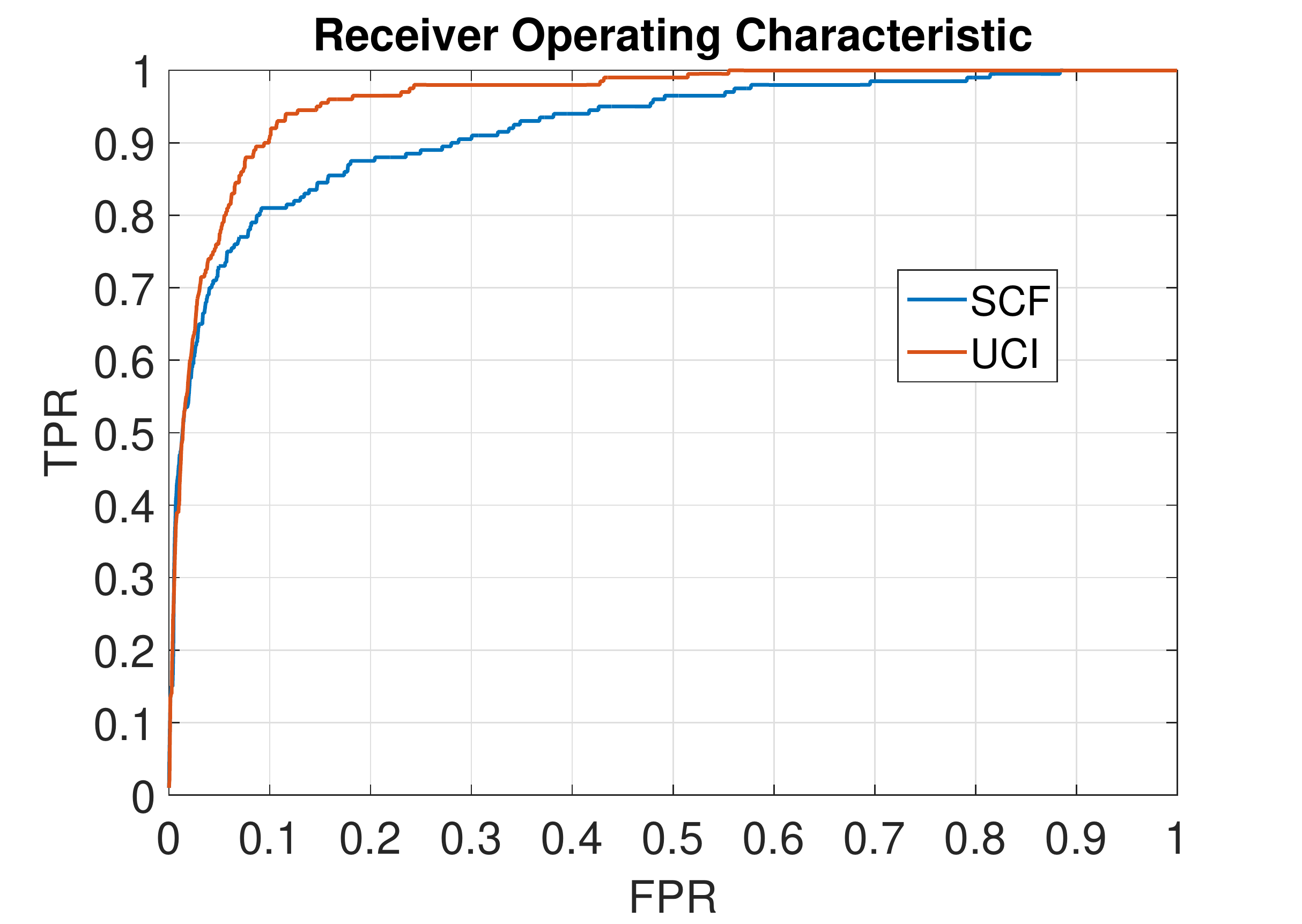}
}
\caption{\small
The ROC curves on: \textbf{(a): Offline}$_1$,\textbf{(b): Offline}$_2$,\textbf{(c): Offline}$_3$  \textbf{(d): Online} dataset, respectively. 
} 
\label{F08_(bar).jpg}
\end{figure*}
That is, the precision should be high. As expected, the proposed method outperforms the baseline in clustering precision in all datasets and varied camera amounts. In brief, using the proposed incremental grouping method, most of images coming from the same camera are returned in the same group, and confusion is significantly alleviated. 

\subsection{Evaluation on Reposted Image Removing}
\begin{table}[h]
\centering  
\caption{The Ratios of Removed Reposted Image Number and False Rejected Image Number}
\vspace{10pt}
\renewcommand\arraystretch{1.5}
\begin{tabular}{lccc|c} 
\hline
& \textbf{Offline}$_1$ & \textbf{Offline}$_2$ & \textbf{Offline}$_3$  & \textbf{Online}\\ 

\hline
Reposted & 49.7\% & 69.0\% & 77.8\%  & 57.8\% \\         
False Rejected  & 15.2\% & 36.6\% & 46.4\%  & 26.8\% \\      
\hline \label{table: 5}
\end{tabular}

\end{table}
In order to alleviate the effect of the reposted images, we perform a post-processing step to automatically remove the possible groups with reposted images. In this step, the groups whose sizes are lower than a predefined threshold $\lambda$ are filtered out. However, in addition to the post-processing step, the incremental clustering procedure also rejects many reposted images. Therefore, we evaluate the effectiveness of reposted image removing by taking both steps into account. In addition to ratio of correctly rejected reposted images, we also make a statistic on false rejected images. The statistical results are shown in Table \ref{table: 5}. As shown in Table \ref{table: 5}, the ratios of correctly rejected reposted images are always remarkably higher than the ratios of false rejected images in all the cases. That is, the proposed method indeed removes reposted images but preserves positive images as well, which can remarkably contribute to the estimation of multiple camera fingerprints. Meanwhile, we can also observe that the ratios for both cases are remarkably increased with the growth of camera number, 
from \textbf{Offline}$_1$ (one camera) to \textbf{Offline}$_3$ (three cameras). In fact, it is reasonable. When the number of cameras increases, the probability that the reposted images are selected as seeds will increased. In this way, more reposted images will be clustered into the same group and be filtered out in group filtering. Meanwhile, more positive images also be rejected, since the error between image and group centroids are more larger.

\subsection{User Identification Evaluation}

In this section, we investigate the performance of our identification schemes. In fact, to the best of our knowledge, it is the first time to re-inspect the user identification problem from the camera fingerprint view. Therefore, no previous works are available for peer comparisons. To fully show the feasibility of the proposed framework, only the schemes designed by us are used for comparisons.

In our experiments, we evaluate the identification performance from two aspects. First, we treat user identification problem as a user retrieval problem. That is, we take each user as a query and retrieve
the users that belong to same individual with the query. We evaluate the performance by the MAP, and the results are listed in Table \ref{table:6}.
\begin{table}[thp]
\centering  
\caption{MAP of Users Identification}
\vspace{10pt}
\renewcommand\arraystretch{1.5}
\begin{tabular}{lccc}  
\hline
& \textbf{SCF} & \textbf{MCF} & \textbf{UCI} \\ 
\hline 
\textbf{Offline$_1$}& 0.3877 & 0.6140 &  \textbf{0.6203} \\      
\textbf{Offline$_2$}& 0.4365 & 0.8201 & \textbf{0.8340} \\
\textbf{Offline$_3$}& 0.8055 & 0.9870 & \textbf{0.9877} \\
\textbf{Online}& 0.7790 & 0.8736 & \textbf{0.9013} \\
\hline \label{table:6}
\end{tabular}

\end{table}

As expected, all the camera fingerprints based user identification methods work well on all datasets. It means that re-inspecting the user identification problem from the camera fingerprint view is quite effective. In addition, after taking into account the multiple camera and image reposting problems, the user identification performance can be further improved clearly. Therefore, it is necessary to handle the multiple camera problem and reposting problem.

Secondly, we also employ ROC curves to further evaluate the performance of the proposed method. The experimental results are illustrated in Fig. \ref{F08_(bar).jpg}. Clearly, for both online and offline datasets, \textbf{UCI} remarkably outperforms the single-clustering method, which is consistent with the MAP.

\section{Related Work}
The problem of user identification has been studied in different research communities \cite{Zhou2016, Ji2015OnKnowledge, Zafarani2015, Liu2015}. In essence, the core of the problem is to find a certain similarity measure to assess the relationships among accounts, \ie, account matching problem. More specifically, it is to extract some discriminative features from accounts so as to change the account identifying problem to the feature matching problem. Therefore, the key of account matching is to extract reliable and effective features of accounts. 

A kind of commonly used method is to extract features from  users' public profiles, such as username, E-mail address, cellphone number. Based on these features, some simple but effective algorithms can be designed to identify multiple accounts across different platforms \cite{Zafarani2013}. Several related works \cite{Perito2011}, \cite{Zafarani2013} have been reported, and good performance on some datasets has been achieved. To further improve the reliability of account matching, more profile attributes (\eg, location \cite{Korayem2013}, interaction activities \cite{Iofciu2011,Goga2015}, friends \cite{Jain2013}) are involved into the account matching process. An assumption underlying these methods is that people maintain the same or similar profiles in different accounts. However, the assumption does not always hold, since people frequently register different accounts with different profiles due to certain purposes (~\eg, privacy concerns, fraud). Therefore, these identification methods cannot deal with such cases. In addition, information barriers among different social network platforms further limit their scope of applications.

To address the issues above, more attentions have been paid on matching accounts by exploiting the user's behaviors in online social networks (\eg, linguistic stylistics, preferred geographic radius, hobbies and interests). The basic assumption is that the same person maintains similar behavior patterns among his/her multiple accounts. For example, some algorithms \cite{Novak,Narayanan2012,Liu2013,Cortis2012} attempt to identify multiple accounts by matching linguistic stylistics of their posted content. In addition, due to the popularization of GPS and cellphone, geo-tagged information posted by online users also become an effective discrimination features \cite{Korayem2013,Malhotra2012}. Although these user identification schemes achieve a good performance on some datasets, they still have some limitations. An obvious drawback is that these methods require massive training samples. Generally, long text and geo-information are always incomplete for most users, which will remarkably influence the identification effectiveness.

Another kind of method is to change user identifying problem into approximate graph isomorphism problem, which can identify more users when a few seed links are given \cite{Korula2014}. In essence, these schemes \cite{Narayanan2012,Zhang2011,Schoenebeck2013,Tan2014,Ji2015OnKnowledge,Goga2014} identify users by matching accounts' networks graph structure, ~\ie, they take users' topology property as discrimination features. This kind of methods is suitable for large scale heterogeneous network reconciliation, however, they are limited to deal with the problem of identifying users on same social network platforms.

In this paper, we attempt to re-inspect the user identification problem from a new perspective,~\ie, camera fingerprint.
Camera fingerprint is a noise-like invisible component existed in digital images, and unique to each imaging equipment \cite{Lukas2006}. With this property, camera fingerprint can be used to determine image's camera source \cite{Chen2008, Bertini2015, Satta2014, Peng2014, Valsesia2015,Caldelli2010} and plenty of related researches have been proposed \cite{Li2010,Kang2012,Thai2014,Tomioka2011,Valsesia2015}.

In fact, \cite{Bertini2015} and \cite{Castiglione2013} are closely related with our work. The former proposes a picture-to-identity linking algorithm to investigate the owner of a particular image, the latter aims to find out the corresponding accounts of a specific camera. However, the problems to be addressed are quite different from the proposed scheme.

\section{Conclusion}

In this paper, we attempt to address the problem of user identification from a new perspective. Instead of using the public information explicitly released by users, we attempt to employ a more reliable feature, i.e., camera fingerprint, to identify multiple accounts belonging to the same individual. To further alleviate the hard problems of multiple cameras and reposting, a novel incremental multi-camera fingerprint estimation algorithm is introduced into the identifying process. The experimental results show that using camera fingerprint information indeed effectively tackles the social media reconciliation problem and the proposed method indeed remarkably alleviates both the multiple cameras and reposting problems.
\bibliographystyle{IEEEtran}
\balance
\bibliography{IEEEabrv,ref}

\begin{thebibliography}{10}
\providecommand{\url}[1]{#1}
\csname url@samestyle\endcsname
\providecommand{\newblock}{\relax}
\providecommand{\bibinfo}[2]{#2}
\providecommand{\BIBentrySTDinterwordspacing}{\spaceskip=0pt\relax}
\providecommand{\BIBentryALTinterwordstretchfactor}{4}
\providecommand{\BIBentryALTinterwordspacing}{\spaceskip=\fontdimen2\font plus
\BIBentryALTinterwordstretchfactor\fontdimen3\font minus
  \fontdimen4\font\relax}
\providecommand{\BIBforeignlanguage}[2]{{%
\expandafter\ifx\csname l@#1\endcsname\relax
\typeout{** WARNING: IEEEtran.bst: No hyphenation pattern has been}%
\typeout{** loaded for the language `#1'. Using the pattern for}%
\typeout{** the default language instead.}%
\else
\language=\csname l@#1\endcsname
\fi
#2}}
\providecommand{\BIBdecl}{\relax}
\BIBdecl

\bibitem{Kemp2015}
S.~Kemp, ``Digital, social \& mobile worldwide in 2015,'' \emph{We Are Social},
  Available:
  http://wearesocial.com/uk/special-reports/digital-social-mobile-worldwide-2015.

\bibitem{Tang}
J.~Tang, C.~Zhang, K.~Cai, L.~Zhang, and Z.~Su, ``{Sampling Representative
  Users from Large Social Networks},'' \emph{Association for the Advancement of
  Artificial Intelligence}, 2015.

\bibitem{Wei2016}
W.~Wei, G.~Cong, C.~Miao, F.~Zhu, and G.~Li, ``{Learning to Find Topic Experts
  in Twitter via Different Relations},'' \emph{IEEE Transactions on Knowledge
  and Data Engineering}, vol.~28, no.~7, pp. 1764--1778, 2016.

\bibitem{Li2016}
J.~Li, C.~Liu, J.~Yu, Y.~Chen, T.~Sellis, and J.~Culpepper, ``{Personalized
  Influential Topic Search via Social Network Summarization},'' \emph{IEEE
  Transactions on Knowledge and Data Engineering}, vol.~28, no.~7, pp.
  1820--1834, 2016.

\bibitem{Li2016a}
H.~Li, Z.~Bu, A.~Li, Z.~Liu, and Y.~Shi, ``{Fast and Accurate Mining the
  Community Structure : Integrating Center Locating and Membership
  Optimization},'' \emph{IEEE Transactions on Knowledge and Data Engineering},
  vol.~28, no.~9, pp. 2349--2362, 2016.

\bibitem{Fang2015}
Q.~Fang, J.~Sang, C.~Xu, and M.~Hossain, ``{Relational User Attribute Inference
  in Social Media},'' \emph{IEEE Transactions on Multimedia}, vol.~17, no.~7,
  pp. 1031--1044, 2015.

\bibitem{Satta2014}
R.~Satta and P.~Stirparo, ``{On the usage of Sensor Pattern Noise for
  Picture-to-Identity Linking through Social Network Accounts},''
  \emph{International Conference on Computer Vision Theory and Applications},
  pp. 5--11, 2014.

\bibitem{Interpreting2014}
S.~Tan, Y.~Li, H.~Sun, Z.~Guan, X.~Yan, J.~Bu, C.~Chen, and X.~He,
  ``{Interpreting the Public Sentiment Variations on Twitter},'' \emph{IEEE
  Transactions on Knowledge and Data Engineering}, no.~5, pp. 1158–--1170,
  2013.

\bibitem{Cross-Platform2016}
X.~Zhou, X.and~Liang, H.~Zhang, and Y.~Ma, ``{Cross-Platform Identification of
  Anonymous Identical Users in Multiple Social Media Networks},'' \emph{IEEE
  transactions on knowledge and data engineering}, no.~2, p. 411–424, 2016.

\bibitem{Personalized2014}
X.~Qian, H.~Feng, G.~Zhao, and T.~Mei, ``{Personalized Recommendation Combining
  User Interest and Social Circle},'' \emph{IEEE Transactions on Multimedia},
  no.~7, pp. 1763–--1777, 2014.

\bibitem{Iofciu2011}
T.~Iofciu, P.~Fankhauser, F.~Abel, and K.~Bischoff, ``{Identifying Users Across
  Social Tagging Systems},'' \emph{International AAAI Conference on Web and
  Social Media}, 2011.

\bibitem{Narayanan2012}
A.~Narayanan, H.~Paskov, G.~{N. Z.}, J.~Bethencourt, E.~Stefanov, E.~Shin, and
  D.~Song, ``{On the Feasibility of Internet-scale Author Identification},''
  \emph{2012 IEEE Symposium on Security and Privacy}, pp. 300--314, 2012.

\bibitem{Zafarani2013}
R.~Zafarani and H.~Liu, ``{Connecting Users across Social Media Sites : A
  Behavioral-Modeling Approach},'' \emph{Proceedings of the 19th ACM SIGKDD
  international conference on Knowledge discovery and data mining. ACM,}, pp.
  41--49, 2013.

\bibitem{Lukas2006}
J.~Luk{\'{a}}{\v{s}}, J.~Fridrich, and M.~Goljan, ``{Digital Camera
  Identification from Sensor Pattern Noise},'' \emph{IEEE Transactions on
  Information Forensics and Security}, vol.~1, no.~2, pp. 205--214, 2006.

\bibitem{Chen2008}
M.~Chen, J.~Fridrich, M.~Goljan, and J.~Luk{\'{a}}{\v{s}}, ``{Determining Image
  Origin and Integrity Using Sensor Noise},'' \emph{IEEE Transactions on
  Information Forensics and Security}, vol.~3, no.~1, pp. 74--90, 2008.

\bibitem{Introduction2008}
D.~Christopher, R.~Prabhakar, and S.~Hinrich, ``{Introduction to Information
  Retrieval},'' \emph{Cambridge University Press}, pp. 356--360, 2008.

\bibitem{Zhou2016}
X.~Zhou, X.~Liang, H.~Zhang, and Y.~Ma, ``{Cross-Platform Identification of
  Anonymous Identical Users in Multiple Social Media Networks},'' \emph{IEEE
  Transactions on Knowledge and Data Engineering}, vol.~28, no.~2, pp.
  411--424, 2016.

\bibitem{Ji2015OnKnowledge}
S.~Ji, W.~Li, N.~Gong, P.~Mittal, and R.~Beyah, ``{On Your Social Network
  De-anonymizablity: Quantification and Large Scale Evaluation with Seed
  Knowledge},'' \emph{NDSS}, pp. 8--11, 2015.

\bibitem{Zafarani2015}
R.~Zafarani and H.~Tang, L.and~Liu, ``{User Identification Across Social
  Media},'' \emph{ACM Transcations on Knowledge Discovery from Data}, vol.~10,
  no.~2, 2015.

\bibitem{Liu2015}
S.~Liu, S.~Wang, and F.~Zhu, ``{Structured Learning from Heterogeneous Behavior
  for Social Identity Linkage},'' \emph{IEEE Transactions on Knowledge and Data
  Engineering}, vol.~27, no.~7, pp. 2005--2019, 2015.

\bibitem{Perito2011}
D.~Perito, C.~Castelluccia, M.~Kaafar, and P.~Manils, ``{How Unique and
  Traceable are Usernames?}'' \emph{International Symposium on Privacy
  Enhancing Technologies Symposium}, pp. 1--17, 2011.

\bibitem{Korayem2013}
M.~Korayem and D.~Crandall, ``{De-anonymizing Users Across Heterogeneous Social
  Computing Platforms},'' \emph{International AAAI Conference on Web and Social
  Media}, 2013.

\bibitem{Goga2015}
O.~Goga, P.~Loiseau, R.~Sommer, R.~Teixeira, and K.~P. Gummadi, ``{On the
  Reliability of Profile Matching Across Large Online Social Networks},''
  \emph{Proceedings of the 21th ACM SIGKDD International Conference on
  Knowledge Discovery and Data Mining. ACM}, pp. 1799--1808, 2015.

\bibitem{Jain2013}
P.~Jain, P.~Kumaraguru, and A.~Joshi, ``{@I Seek 'Fb.Me': Identifying Users
  Across Multiple Online Social Networks},'' \emph{Proceedings of the 22nd
  International Conference on World Wide Web}, pp. 1259--1268, 2013.

\bibitem{Novak}
J.~Novak and A.~Tomkins, ``{Anti-Aliasing on the Web},'' \emph{Proceedings of
  the 13th International Conference on World Wide Web. ACM}, pp. 30--39, 2004.

\bibitem{Liu2013}
J.~Liu, F.~Zhang, X.~Song, Y.~Song, C.~Lin, and H.~Hon, ``{What's In A Name?:
  An Unsupervised Approach to Link Users Across Communities},''
  \emph{Proceedings of the Sixth ACM International Conference on Web Search and
  Data Mining}, pp. 495--504, 2013.

\bibitem{Cortis2012}
K.~Cortis, S.~Scerri, I.~Rivera, and S.~Handschuh, ``{Discovering Semantic
  Equivalence of People Behind Online Profiles},'' \emph{Proceedings of the
  Resource Discovery (RED) Workshop.}, 2012.

\bibitem{Malhotra2012}
A.~Malhotra, L.~Totti, W.~Meira, P.~Kumaraguru, and V.~Almeida, ``{Studying
  User Footprints in Different Online Social Networks},'' \emph{Proceedings of
  the 2012 International Conference on Advances in Social Networks Analysis and
  Mining}, pp. 1065--1070, 2012.

\bibitem{Korula2014}
N.~Korula and S.~Lattanzi, ``{An Efficient Reconciliation Algorithm for Social
  Networks},'' \emph{Proceedings of the VLDB Endowment}, vol.~7, no.~5, pp.
  377--388, 2014.

\bibitem{Zhang2011}
Y.~Zhang, J.~Tang, Z.~Yang, J.~Pei, and P.~Yu, ``{COSNET : Connecting
  Heterogeneous Social Networks with Local and Global Consistency Categories
  and Subject Descriptors},'' \emph{Proceedings of the 21th ACM SIGKDD
  International Conference on Knowledge Discovery and Data Mining}, pp.
  1485--1494, 2011.

\bibitem{Schoenebeck2013}
G.~Schoenebeck, ``{Potential Networks, Contagious Communities, and
  Understanding Social Network Structure},'' \emph{International Conference on
  World Wide Web, ACM}, pp. 1123--1132, 2013.

\bibitem{Tan2014}
S.~Tan, Z.~Guan, D.~Cai, X.~Qin, J.~Bu, and C.~Chen, ``{Mapping Users across
  Networks by Manifold Alignment on Hypergraph},'' \emph{Association for the
  Advancement of Artificial Intelligence}, pp. 159--165, 2014.

\bibitem{Goga2014}
O.~Goga, ``{Matching User Accounts Across Online Social Networks: Methods and
  Applications},'' \emph{Doctoral dissertation, LIP6-Laboratoire d'Informatique
  de Paris 6}, 2014.

\bibitem{Bertini2015}
F.~Bertini, R.~Sharma, A.~Iann{\`{\i}}, and D.~Montesi, ``{Profile Resolution
  across Multilayer Networks through Smartphone Camera Fingerprint},''
  \emph{Proceedings of the 19th International Database Engineering \&
  Applications Symposium, ACM}, pp. 23--32, 2015.

\bibitem{Peng2014}
F.~Peng, J.~Shi, and M.~Long, ``{Identifying Photographic Images and
  Photorealistic Computer Graphics Using Multifractal Spectrum Features of
  PRNU},'' \emph{IEEE International Conference on Multimedia and Expo}, pp.
  1--6, 2014.

\bibitem{Valsesia2015}
D.~Valsesia, G.~Coluccia, T.~Bianchi, and E.~Magli, ``{Large-scale Image
  Retrieval Based on Compressed Camera Identification},'' \emph{IEEE
  Transactions on Multimedia.}, vol.~17, no.~9, pp. 1439--1449, 2015.

\bibitem{Caldelli2010}
R.~Caldelli, I.~Amerini, F.~Picchioni, and M.~Innocenti, ``{Fast Image
  Clustering of Unknown Source Images},'' \emph{IEEE International Workshop on
  Information Forensics and Security}, pp. 1--5, 2010.

\bibitem{Li2010}
L.~Chang-tsun, ``{Source Camera Identification Using Enhanced Sensor Pattern
  Noise},'' \emph{IEEE Transactions on Information Forensics and Security},
  vol.~5, no.~2, pp. 280--287, 2010.

\bibitem{Kang2012}
X.~Kang, Y.~Li, Z.~Qu, and J.~Huang, ``{Enhancing Source Camera Identification
  Performance with a Camera Reference Phase Sensor Pattern Noise},'' \emph{IEEE
  Transactions on Information Forensics and Security}, vol.~7, no.~2, pp.
  393--402, 2012.

\bibitem{Thai2014}
T.~Thai, R.~Cogranne, and F.~Retraint, ``{Camera Model Identification Based on
  the Heteroscedastic Noise Model},'' \emph{IEEE Transactions on Image
  Processing.}, vol.~23, no.~1, pp. 250--263, 2014.

\bibitem{Tomioka2011}
Y.~Tomioka and H.~Kitazawa, ``{Digital Camera Identification Based on the
  Clustered Pattern Noise of Image Sensors},'' \emph{IEEE International
  Conference on Multimedia and Expo}, pp. 1--4, 2011.

\bibitem{Castiglione2013}
A.~Castiglione, G.~Cattaneo, M.~Cembalo, and U.~Petrillo, ``{Experimentations
  With Source Camera Identification and Online Social Networks},''
  \emph{Journal of Ambient Intelligence and Humanized Computing}, vol.~4,
  no.~2, pp. 265--274, 2013.

\end{thebibliography}
\end{document}